\documentclass{article}
\PassOptionsToPackage{numbers,compress}{natbib}
\usepackage[main, preprint]{neurips_2026}
\usepackage[utf8]{inputenc}
\usepackage[T1]{fontenc}
\usepackage{microtype}
\usepackage{graphicx}
\usepackage{enumitem}
\usepackage{subcaption}
\usepackage{booktabs}
\usepackage{hyperref}
\usepackage{url}
\usepackage{wrapfig}
\usepackage{xcolor}
\usepackage{amsfonts}
\usepackage{nicefrac}
\usepackage{amsmath, amssymb, mathtools, amsthm}
\usepackage{algorithm}
\usepackage{algorithmic}
\usepackage{float}
\usepackage[capitalize,noabbrev]{cleveref}
\theoremstyle{plain}
\usepackage{bbding}
\newtheorem{theorem}{Theorem}[section]
\newtheorem{proposition}[theorem]{Proposition}

\theoremstyle{definition}

\theoremstyle{remark}

\title{Stability-Aware Self-Training for CLIP under Cross-Modal Anchoring Mismatch}
\author{
  Siyuan Liu\textsuperscript{1},
  Xinyang Chen\textsuperscript{1}\Envelope,
  Xiucheng Li\textsuperscript{1},
  Weili Guan\textsuperscript{2},
  Liqiang Nie\textsuperscript{1} \\
  \textsuperscript{1}School of Computer Science and Technology,
  Harbin Institute of Technology (Shenzhen) \\
  \textsuperscript{2}School of Information Science and Technology,
  Harbin Institute of Technology (Shenzhen) \\
  \texttt{\{liusiyuan666777, chenxinyang95, nieliqiang\}@gmail.com,}\\
  \texttt{\{lixiucheng, guanweili\}@hit.edu.cn}
}
\begin{document}
\maketitle

\begin{abstract}
{\parfillskip=0pt\par
Self-training is a practical way to adapt CLIP under limited supervision, but its performance is highly sensitive to pseudo-label reliability. We identify a key failure mode in CLIP adaptation: temporal prediction instability, in which a target sample repeatedly switches its prediction across epochs. We offer a perspective based on cross-modal anchoring mismatch: the decision geometry induced by pre-trained text anchors can deviate from the local visual cluster structure of target domain. This mismatch can make ambiguous target samples susceptible to prediction switches during adaptation, whereas target-aligned anchoring can produce more stable pseudo-labels.
Motivated by this observation, we propose Stability-Aware Self-Training (\textbf{SAST}), centered on Stability-Weighted Image Prototypes (\textbf{SWIP}). SWIP builds class-wise visual anchors from temporally stable target samples to provide more reliable adaptation references. We further add a lightweight text-side refinement to reduce confusion-induced instability and combine the two streams with an adaptive fusion rule. Across six benchmarks and three adaptation settings, SAST consistently outperforms strong baselines. The results suggest that explicitly modeling temporal stability is an effective route to robust CLIP adaptation under self-training. Our code is available at \url{https://github.com/lsy-2003/SAST}.
\par}
\end{abstract}

{\parfillskip=0pt\par
\section{Introduction}
\label{sec:intro}
Large-scale vision-language models (VLMs)~\citep{radford2021learning, li2021align, zhai2022lit} have shown strong transferability across a wide range of visual recognition tasks. To adapt these models to downstream domains with low annotation cost, self-training has become a practical paradigm that iteratively exploits unlabeled target data through pseudo-label supervision~\citep{sohn2020fixmatch, zhou2022conditional, gao2024clip}. Its effectiveness depends critically on pseudo-label reliability~\citep{sohn2020fixmatch, chen2023softmatch, wang2025handling}. In CLIP~\citep{radford2021learning} adaptation, however, this reliability can be fragile: predictions that appear confident at one epoch may change later, causing pseudo-label to fluctuate during training.

In this work, we focus on an important yet less-studied failure mode in CLIP self-training: prediction instability, i.e., changes in predicted labels across epochs. Such instability is harmful because inconsistent predictions can be repeatedly reused as pseudo-labels, injecting unreliable supervision into subsequent adaptation~\citep{sohn2020fixmatch, he2024gradual}. Existing methods typically evaluate pseudo-label reliability at a single point during training, by regulating the number of pseudo-labels, the sharpness of confidence, or calibrating probabilities~\citep{chen2023softmatch, guo2017calibration}. However, a pseudo-label that is confident at one epoch can still be unreliable if its prediction is temporally unstable. This motivates our view that stability should not only determine which pseudo-labels to trust, but what targets the model should adapt toward.

Our study is motivated by the modality gap~\citep{liang2022mind, schrodi2024two}, which shows that image and text embeddings in vision-language models can exhibit systematic separation. In this paper, we focus on a related but distinct issue during adaptation, and we refer to this as \textbf{Cross-Modal Anchoring Mismatch}: pre-trained text anchors (produced by the text encoder) may not align well with the target-domain visual clusters. Although text anchors remain valuable because they provide zero-shot semantic knowledge, their local classification boundaries may still be imperfect for the target domain. When a target sample lies near the boundary induced by misaligned text anchors, small changes in the adapted representation can frequently change its nearest class anchor, leading to temporal prediction switches.

To explore how the arrangement of class anchors is related to prediction stability, we compare the training dynamics of three variants under the 2-shot SSL setting: a standard adaptation based on text anchors, a target-aligned visual anchor reference (initialized by visual features of the 2-shot samples), and our SAST framework. We provide further analysis and interpretation in Appendices~\ref{app:instability_analysis} and~\ref{app:interpretation}.

\begin{figure}[t]
    \centering
    \begin{subfigure}[b]{0.48\columnwidth} 
        \centering
        \includegraphics[width=\linewidth]{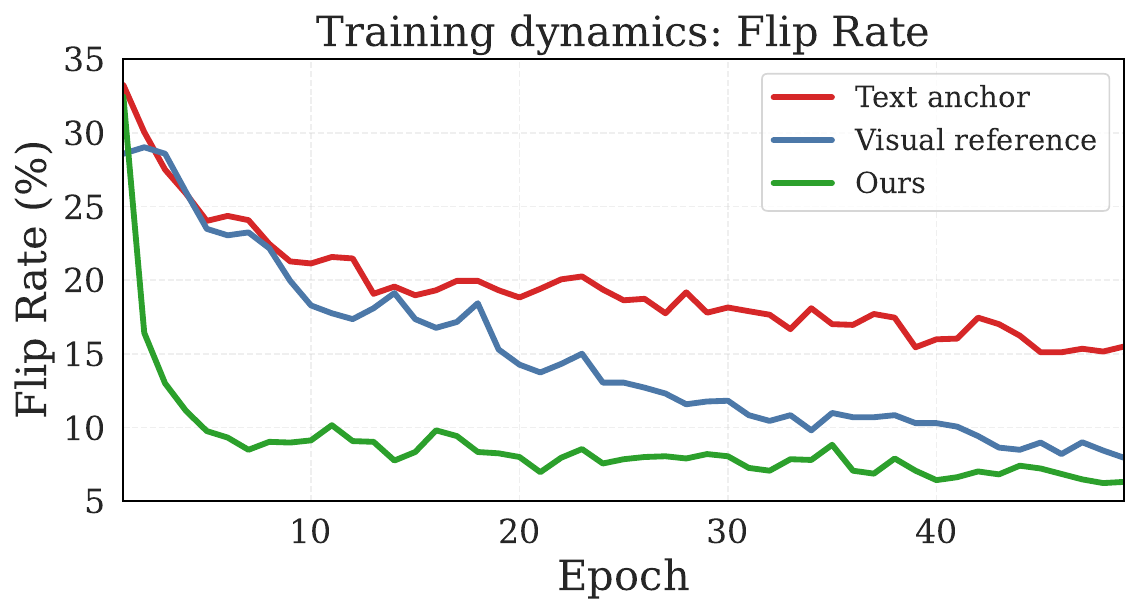}
        \caption{Switch rate.}
        \label{fig:dy1}
    \end{subfigure}
    \hspace{1em} 
    \begin{subfigure}[b]{0.48\columnwidth}
        \centering
        \includegraphics[width=\linewidth]{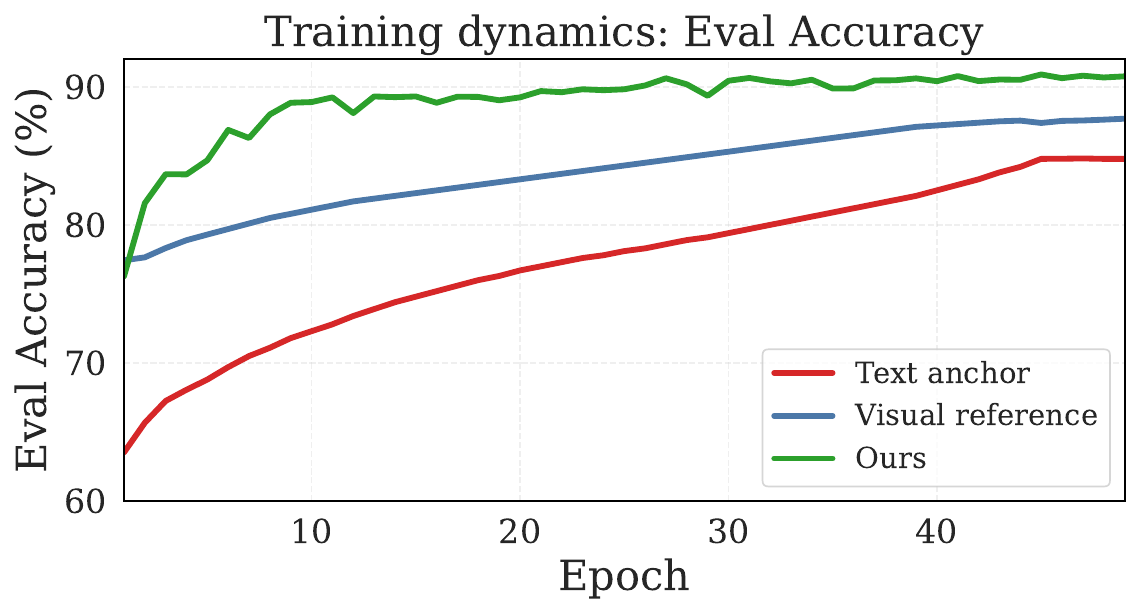}
        \caption{Accuracy.}
        \label{fig:dy2}
    \end{subfigure}
    \caption{Training dynamics under SSL setting. We compare three variants: a text anchor adaptation, a target-aligned visual anchor reference and our SAST. For detailed protocol, refer to Appendix~\ref{app:diag_protocol}.}
    \label{fig:intro_dynamics}
\end{figure}

As shown in Figure~\ref{fig:intro_dynamics}, the target-aligned visual reference exhibits consistently lower switch rates and higher accuracy throughout the self-training process. The resulting prediction trajectories become more stable, suggesting that instability is not simply caused by insufficient training, but is instead closely related to the anchors used during adaptation. Moreover, the stability gain persists across the entire adaptation process, indicating that target-aligned anchoring affects not only the initial decision geometry but also the subsequent pseudo-label dynamics. This observation motivates a central design principle of our method: instead of relying only on fixed text anchors, CLIP adaptation can benefit substantially from target-aligned visual references that are constructed from reliable target samples.

Therefore, we propose \textbf{Stability-Aware Self-Training (SAST)}, which uses temporal stability as a structural signal to construct class anchors and to guide fusion. The core component, \textbf{Stability-Weighted Image Prototypes (SWIP)}, builds class-wise visual anchors from temporally stable target samples. To further reduce instability among semantically nearby classes, we introduce \textbf{Confusion-Aware Text Prompt Augmentation (CTPA)} and combine these two streams with an adaptive fusion rule. In Figure~\ref{fig:intro_dynamics}, SAST achieves the lowest switch rate and the highest accuracy among the three variants, supporting our hypothesis that stability-aware visual anchoring can effectively mitigate prediction instability. We evaluate SAST on CLIP adaptation benchmarks that emphasize domain shift and fine-grained ambiguity, where pseudo-label reliability remains an important bottleneck.

Our contributions are summarized as follows. First, we identify \textbf{prediction instability} as an important yet underexplored failure mode in CLIP self-training and provide diagnostic evidence that associates it with anchoring mismatch in adaptation. Second, we propose \textbf{SWIP}, which constructs class-wise visual anchors from temporally stable samples to serve as target-aligned references during self-training. Third, based on SWIP and a lightweight text refinement, we develop \textbf{SAST}, a stability-aware adaptation framework, and show consistent gains across multiple CLIP adaptation settings.

\section{Related Work}

\textbf{Pseudo-label-based CLIP adaptation.} Recent CLIP adaptation methods increasingly exploit unlabeled target data through pseudo-label self-training~\cite{huang2022unsupervised, menghini2023enhancing, zhang2024candidate, wang2025handling}. Representative approaches improve pseudo-label usage via confidence-based sample selection, candidate-label modeling, or class balancing. These are the most direct baselines to our work, since they operate under the same practical setting of adapting CLIP with unlabeled or weakly labeled target data. In particular, we focus on a specific and underexplored failure mode in this regime: pseudo-labels that remain temporally unstable across training, which can inject unreliable supervision and therefore degrade adaptation.

\textbf{Pseudo-label reliability and consistency.}
In semi-supervised learning, pseudo-label reliability is commonly improved through techniques such as confidence thresholding, calibration, consistency regularization, and curriculum strategies~\cite{laine2016temporal, sohn2020fixmatch, zhang2021flexmatch, chen2023softmatch, guo2017calibration, cascante2021curriculum, wang2022freematch}. Related analyses of self-training dynamics also highlight the accumulation of pseudo-label errors during adaptation~\cite{he2024gradual}. Our setting differs in two important respects. First, while consistency regularization in prior SSL work mainly targets perturbation-level invariance, we focus instead on epoch-level prediction switches during CLIP adaptation under anchoring mismatch. Second, temporal stability in SAST serves as a structural signal that guides visual anchor construction, unlabeled-sample partitioning, and text--visual fusion.

\textbf{Prompt and cache-based CLIP adaptation.}
Prompt tuning, adapter tuning, and cache/prototype methods have been widely studied for CLIP adaptation with labeled support sets~\cite{zhou2022learning, zhou2022conditional, khattak2023maple, lee2023read, jia2022visual, bahng2022exploring, zhang2022tip}. Related cache-based or training-free methods, such as Tip-Adapter~\cite{zhang2022tip} and TFUP~\cite{long2024training}, also improve adaptation by constructing representative memories or prompts from target data. SWIP repeatedly identifies temporally stable target samples and uses them to refresh class-wise visual anchors over the course of self-training, instead of depending on support prototypes or static caches.

\textbf{Attribute-guided prompting.}
Several recent methods enrich CLIP text representations by incorporating semantic attributes, class descriptions, or confusion-aware prompting strategies~\cite{wang2025handling, cai2025attribute, tian2024argue, shao2026capt}. In contrast, our text-side refinement component is considerably lighter-weight: CTPA performs a fixed offline augmentation and selects neighbor-discriminative attributes only for confusion-prone classes, rather than relying on end-to-end prompt tuning or more complex attribute fusion schemes.

\begin{figure*}[t]
\begin{center}
\centerline{\includegraphics[width=0.9\textwidth]{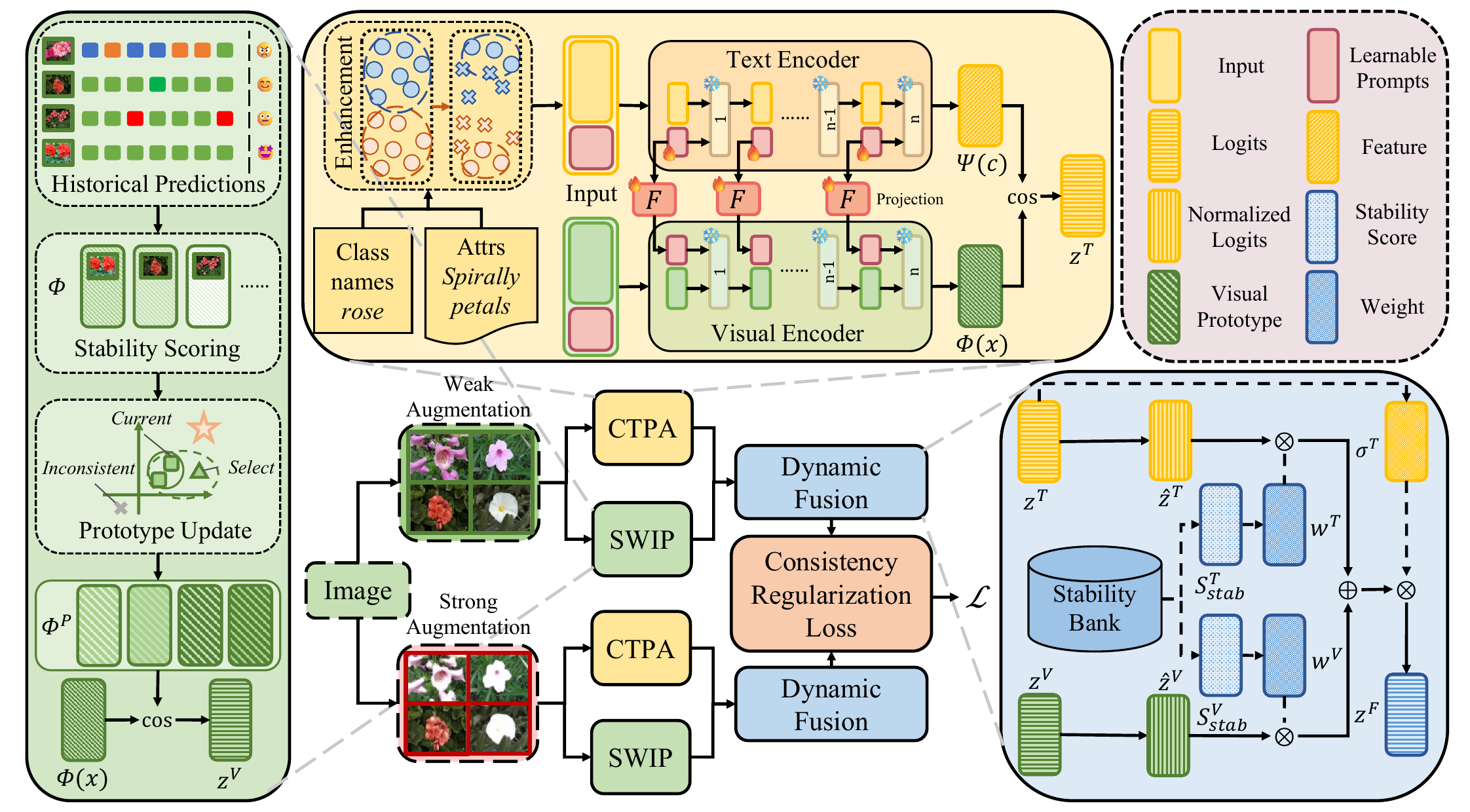}} 
\caption{Our SAST framework. 
\textbf{\textcolor[HTML]{618E45}{Green:}} SWIP selects class-wise image prototypes from temporally stable samples to provide visual anchors.  
\textbf{\textcolor[HTML]{FFC925}{Yellow:}} CTPA augments confusion-prone class prompts with discriminative attributes to enlarge local text-side separation.  
\textbf{\textcolor[HTML]{466BA4}{Blue:}} The rule to obtain a fused prediction. The CLIP encoders are frozen; only the learnable prompts (following MaPLe) are trained.} 
\label{fig:framework}
\vspace*{-10pt}
\end{center}
\end{figure*}

\section{Methodology}
\label{sec:method}

\textbf{Problem Definition.}
We adapt a pre-trained CLIP model to a downstream domain with label space $\mathcal{Y}=\{1,\dots,C\}$. The training data consist of a large unlabeled set $\mathcal{D}_{\mathrm{UL}}=\{\mathbf{x}_i\}_{i=1}^N$ and, optionally, a small labeled set $\mathcal{D}_{\mathrm{L}}=\{(\mathbf{x}_i,y_i)\}_{i=1}^M$. Let $\phi(\cdot;\boldsymbol{\theta})$ and $\psi(\cdot;\boldsymbol{\theta})$ denote the visual and text encoders, respectively. Given an image $\mathbf{x}$ and a class prompt $\text{prompt}_c$, CLIP then predicts by cosine similarity:
\[
p(y=c\mid \mathbf{x}) \propto \exp\!\left(\cos\big(\phi(\mathbf{x};\boldsymbol{\theta}),\, \psi(\text{prompt}_c;\boldsymbol{\theta})\big)/\tau\right).
\]
We adapt the model using labeled data when available, and unlabeled data otherwise, with pseudo-labels generated during training. Following CAP~\cite{wang2025handling}, we use MaPLe~\cite{khattak2023maple} as the adaptation scheme.

\textbf{Overview.}
SAST is designed to improve CLIP self-training stability. It centers on SWIP (\S\ref{sec:SWIP}), is complemented by CTPA (\S\ref{sec:ctpa}) and combines the two streams through an adaptive fusion rule (\S\ref{sec:fusion}).

\subsection{Stability-Weighted Image Prototypes}
\label{sec:SWIP}

\textbf{Motivation.}
SWIP mitigates sample-level prediction instability by constructing class-wise visual anchors directly from target data. Since support-derived prototypes are unavailable in fully unlabeled settings, we build them in a self-training manner from temporally stable samples. This yields anchors that better align with the underlying target distribution and are more reliable during self-training.

\textbf{Visual prototype classifier.}
Given an image $\mathbf{x}$, we use its $\ell_2$-normalized visual embedding
$\tilde{\phi}(\mathbf{x})=\phi(\mathbf{x})/\|\phi(\mathbf{x})\|_2$.
Let $\boldsymbol{\Phi}^P=[\boldsymbol{\mu}_1,\dots,\boldsymbol{\mu}_C]$ denote the normalized prototypes. SWIP produces logits by
\begin{equation}
\mathbf{z}^{V}=\tilde{\phi}(\mathbf{x})(\boldsymbol{\Phi}^{P})^\top/\tau.
\end{equation}
Since $\boldsymbol{\Phi}^P$ are refreshed from temporally stable target-domain samples, they provide more reliable visual references that are typically less sensitive to transient prediction switches than text anchors.

\textbf{Stability-based prototype construction.}
We maintain a memory bank $\mathcal{Q}$ that stores prediction history and lightweight statistics. At the beginning of each epoch, SWIP refreshes the prototypes by selecting candidate samples with high temporal stability. For a sample $\mathbf{x}$, let $\{\hat{y}^{(l)}\}_{l=1}^{L}$ denote the available historical predicted classes of $\mathbf{x}$ stored in $\mathcal{Q}$ up to the last epoch $L$, and let $\hat{y}^{(L+1)}$ be its current prediction. We estimate its prediction instability by computing the disagreement frequency
\begin{equation}
S_{\mathrm{inst}}(\mathbf{x})=\frac{1}{L}\sum_{l=1}^{L}\mathbb{I}\!\left(\hat{y}^{(l)}\neq \hat{y}^{(L+1)}\right).
\end{equation}
For samples with insufficient history ($L<2$), we assign a neutral frequency $S_0 = 0.5$ (an ablation is provided in Table~\ref{tab:hyperparam_ablation}). Then the frequency is converted into a stability score by a normalized ranking,
\begin{equation}
S(\mathbf{x}) =
\mathcal{R}\!\left(
1 - S_{\mathrm{inst}}(\mathbf{x})\right)
- \lambda_c\Delta_{\mathrm{cons}}(\mathbf{x})
,
\end{equation}
where $\mathcal{R}(\cdot)$ maps the scores in the current candidate pool to $[0,1]$. Here, $\Delta_{\mathrm{cons}}$ discourages prototype drift away from the text-side anchor. For a candidate $\mathbf{x}$ with pseudo-label $c$, we form a tentative update $\boldsymbol{\mu}_{\mathrm{new}} = \mathrm{Normalize}(\boldsymbol{\mu}_c^{(t)} + \tilde\phi(\mathbf{x}))$ and measure its similarity drop to the text prototype $\boldsymbol{\psi}_c$:
\begin{equation}
\Delta_{\mathrm{cons}}(\mathbf{x})
=
\max\!\left(
0,\,
\cos(\boldsymbol{\mu}_c^{(t)}, \boldsymbol{\psi}_c)
-
\cos(\boldsymbol{\mu}_{\mathrm{new}}, \boldsymbol{\psi}_c)
\right).
\end{equation}

Samples with high $S(\mathbf{x})$ are less affected by the prediction switches and are therefore more suitable to serve as visual anchors. For each class $c$, we collect unlabeled samples whose current pseudo-label is $c$. To avoid class imbalance and reduce computational cost, the selection is performed class-wise. Denoting the selected unlabeled set by $\mathcal{S}_{\mathrm{UL},c}^{(t)}$ and the labeled set by $\mathcal{D}_{\mathrm{L},c}$, we update the prototype as
\begin{equation}
\boldsymbol{\mu}_c^{(t+1)}=
\mathrm{Normalize}\!\left(
\frac{1}{|\mathcal{S}_c^{(t)}|}
\sum_{\mathbf{x}\in\mathcal{S}_c^{(t)}} \tilde\phi(\mathbf{x})
\right),
\end{equation}
where $\mathcal{S}_c^{(t)}=\mathcal{D}_{\mathrm{L},c}\cup \mathcal{S}_{\mathrm{UL},c}^{(t)}$ contains labeled samples of class $c$ (when such labels are available) together with the selected unlabeled target samples. We update the prototypes progressively, starting from highly stable core samples and gradually broadening the candidate pool as training proceeds.

In this way, SWIP leverages temporal stability as the selection criterion to progressively convert unlabeled target samples into increasingly reliable visual anchors over the course of self-training.

\subsection{Confusion-Aware Text Prompt Augmentation}
\label{sec:ctpa}

\textbf{Motivation.}
Beyond sample-level instability, prediction switches concentrate on semantically confusing classes (Appendix~\ref{app:geometry_flip}), suggesting that standard text prompts may insufficiently separate nearby categories. CTPA therefore refines prompts for those classes to enlarge the local text separation.

\textbf{Confusion-aware prompt augmentation.}
We first define the confusion neighbor set of class $c$ as
\begin{equation}
\mathcal{N}(c)=\{c' \in \mathcal{Y}\mid c'\neq c,\ \cos(\psi_{\mathrm{base}}(c),\psi_{\mathrm{base}}(c'))>\delta\},
\end{equation}
where $\psi_{\mathrm{base}}(c)$ is the text feature of the base class name. For each class, an LLM generates a candidate attribute pool $\mathcal{A}_c$ offline, after which the prompts remain fixed. We then rank attributes by
\begin{equation}
\rho(a;c)=\max_{c' \in \mathcal{N}(c)} \cos(\psi(a),\psi_{\mathrm{base}}(c')),
\end{equation}
and select the attributes with the lowest values that are least similar to the confusing neighbors. We construct the retained subset $\mathcal{A}_c^{*}$ greedily: starting from the base class prompt, we add the candidate attribute with the lowest $\rho(a;c)$ among the remaining candidates at each step and stop once the prompt length limit is reached. For classes without confusion neighbors, we keep the prompt unchanged.

The enhanced text feature is formed by concatenating the class name with the selected attributes in $\mathcal{A}_c^{*}$. Stacking the resulting class-wise features yields $\boldsymbol{\Psi}\in\mathbb{R}^{C\times D}$. The corresponding text logits are
\begin{equation}
\mathbf{z}^{T}=\phi(\mathbf{x})\boldsymbol{\Psi}^{\top}/\tau.
\end{equation}
By selectively injecting such confusion-aware attributes, CTPA reduces instability from the textual side, complementing SWIP's visual anchoring. Attributes for all classes are generated offline from class names only, without access to any target samples, labels, or statistics, matching the information available to the baselines. Details of LLM usage with more experiments are provided in Appendix~\ref{app:ctpa_llm}.
\subsection{Prediction Rule and Optimization}
\label{sec:fusion}

\textbf{Prediction Rule.}
Since SAST includes both visual and text-based predictors, we use an adaptive fusion rule to combine their outputs. This adaptivity is driven by sample-level branch reliability. For each stream $\mathrm{Str}\in\{T,V\}$, we define a nonnegative reliability score $R^{\mathrm{Str}}(\mathbf{x})$. During training, recurring unlabeled target samples provide prediction histories over successive epochs, so $R^{\mathrm{Str}}(\mathbf{x}) = 1 - S_{\mathrm{inst}}(\mathbf{x})$ is computed analogously to SWIP from the disagreement frequency between the current branch prediction and its branch-specific prediction history accumulated up to the preceding epoch. At evaluation time, histories are unavailable, so we use branch confidence as an inductive surrogate.

Due to the scale discrepancy between the two streams, we normalize their logits: $\hat{\mathbf{z}}^{\mathrm{Str}} = (\mathbf{z}^{\mathrm{Str}} - \mu^{\mathrm{Str}}) / \sigma^{\mathrm{Str}}$, where $\mu^{\mathrm{Str}}$ and $\sigma^{\mathrm{Str}}$ are the instance mean and standard deviation. We then convert the reliability scores into normalized fusion weights $w^{\mathrm{Str}} \propto R^{\mathrm{Str}}$, and rescale the fused logits to $\sigma^{T}$ so that they remain on the same scale as the native CLIP logits. The fused logits are computed as
\begin{equation}
\mathbf{z}^{F}
=
\sigma^{T}
\Big(
w^{T}\,\hat{\mathbf{z}}^{T}
+
w^{V}\,\hat{\mathbf{z}}^{V}
\Big).
\end{equation}

The resulting prediction rule thus preserves the complementary benefits of SWIP and CTPA while reducing the risk that a less reliable stream dominates the final prediction for a particular sample.

\textbf{Optimization.}
The optimization follows standard pseudo-label training. We optimize the model with
\begin{equation}
\mathcal{L}
=
\mathcal{L}_{\mathrm{sup}}(\mathcal{D}_{\mathrm{L}})
+
\mathcal{L}_{\mathrm{unsup}}(\mathcal{U}),
\end{equation}
where $\mathcal{L}_{\mathrm{sup}}$ is the standard cross-entropy loss. For the unlabeled batch $\mathcal{U}$, we use a standard consistency regularization with CE loss. During optimization, for an unlabeled sample $\mathbf{x}$, we compute the fused prediction $\mathbf{z}^{F}(\mathbf{x})$ and obtain its maximum probability $p_{\max}(\mathbf{x})$. We maintain an EMA-updated confidence threshold $\mu_{\mathrm{conf}}^{(t)}$ initialized at  $\mu_{\mathrm{init}}$. The training batch is formed as:
\begin{equation}
\mathcal{U}^{(t)}
=
\left\{
\mathbf{x}_i \in \mathcal{D}_{\mathrm{UL}}
:
p_{\max}(\mathbf{x}_i) \ge \mu_{\mathrm{conf}}^{(t)}
\right\}.
\end{equation}

\section{Experiments}
\label{sec:experiments}

We evaluate SAST in terms of overall adaptation performance and prediction-instability control. We describe the settings (\S\ref{sec:settings}), report the main results (\S\ref{sec:main_results}), and provide ablations and analyses (\S\ref{sec:analysis}).

\subsection{Experimental Settings}
\label{sec:settings}

\textbf{Datasets and Metrics.}
Following previous studies \cite{menghini2023enhancing, zhang2024candidate, wang2025handling}, we evaluate on the same six widely used benchmarks: RESISC45 \cite{cheng2017remote}, DTD \cite{cimpoi2014describing}, EuroSAT \cite{helber2019eurosat}, FGVC-Aircraft \cite{maji2013fine}, CUB \cite{wah2011caltech}, and Flowers102 \cite{nilsback2008automated}. These datasets encompass diverse visual domains, ranging from satellite imagery and textures to fine-grained object recognition, and are standard in recent pseudo-label-based CLIP adaptation work, where pseudo-label reliability under target-domain mismatch is the central concern. We report the top-1 accuracy averaged across three independent runs with different random seeds.

\textbf{Learning Paradigms.}
We evaluate SAST under three target adaptation settings: 
(1) \textbf{Unsupervised Learning (UL):} adaptation uses only unlabeled target data $\mathcal{D}_{\mathrm{UL}}$. 
(2) \textbf{Semi-Supervised Learning (SSL):} following CAP~\cite{wang2025handling}, we provide two labeled samples per class together with the remaining unlabeled target data. 
(3) \textbf{Transductive Zero-Shot Learning (TRZSL):} following prior work~\cite{menghini2023enhancing, zhang2024candidate, wang2025handling}, we split the label space into 62\% seen and 38\% unseen classes. During adaptation, seen classes are trained with full supervision, whereas unseen classes rely solely on unlabeled target data. We report generalized performance by the harmonic mean of top-1 accuracy on the seen and unseen splits. Here, "transductive" is used in the sense of prior work and does not imply that test samples are used for training. It only refers to the availability of unlabeled samples from unseen classes during adaptation, which yields an imbalanced supervision setting. Evaluation remains inductive at test time.

\textbf{Implementation.}
We use pre-trained CLIP (ViT-B/32) \cite{radford2021learning} as the backbone. The model is trained for 50 epochs with SGD (learning rate of $0.02$, momentum of $0.9$) and a cosine annealing scheduler. All experiments are conducted on an NVIDIA A100 GPU. Training details are provided in Appendix~\ref{app:settings}.

\textbf{Baselines.}
We compare SAST with zero-shot CLIP and several recent CLIP adaptation methods under the same pseudo-label-based target adaptation regime, including FPL/GRIP~\cite{menghini2023enhancing} (NeurIPS 2023), CPL~\cite{zhang2024candidate} (ICML 2024), and CAP~\cite{wang2025handling} (ICML 2025). These are our most direct baselines, since they likewise rely on unlabeled or weakly labeled target data and are similarly affected by pseudo-label reliability under domain shift. By contrast, few-shot methods built on labeled support sets (e.g., prompt-, adapter-, or cache-based tuning) are not protocol-matched and are excluded.

\begin{table*}[t]
\caption{Comparison of accuracy (\%). Results are reported as $\text{mean}_\text{std}$. The best results are in \textbf{bold}.}
\label{tab:main_results}
\begin{center}
\begin{small}
\begin{sc}
\resizebox{\textwidth}{!}{
\begin{tabular}{lccccccccc}
\toprule
 & \multicolumn{3}{c}{Flowers102} & \multicolumn{3}{c}{RESISC45} & \multicolumn{3}{c}{DTD} \\
\cmidrule(lr){2-4} \cmidrule(lr){5-7} \cmidrule(lr){8-10}
Methods & UL & SSL & TRZSL & UL & SSL & TRZSL & UL & SSL & TRZSL \\
\midrule
CLIP \cite{radford2021learning}
    & $63.67_{0.00}$ & $63.67_{0.00}$ & $63.40_{0.00}$
    & $54.48_{0.00}$ & $54.48_{0.00}$ & $54.46_{0.00}$
    & $43.24_{0.00}$ & $43.24_{0.00}$ & $43.45_{0.00}$ \\
FPL \cite{menghini2023enhancing}
    & $65.67_{0.23}$ & $75.96_{0.74}$ & $80.97_{0.00}$
    & $63.07_{0.38}$ & $68.13_{0.55}$ & $72.11_{0.00}$
    & $44.96_{0.55}$ & $37.10_{5.45}$ & $46.30_{0.03}$ \\
GRIP \cite{menghini2023enhancing}
    & $69.84_{1.06}$ & $83.60_{0.48}$ & $86.26_{0.00}$
    & $70.55_{0.88}$ & $74.11_{0.68}$ & $81.07_{0.00}$
    & $46.09_{1.06}$ & $56.07_{0.85}$ & $65.30_{0.01}$ \\
CPL \cite{zhang2024candidate}
    & $72.90_{0.78}$ & $89.66_{0.36}$ & $87.35_{0.76}$
    & $77.39_{0.44}$ & $80.98_{0.11}$ & $85.85_{0.49}$
    & $51.91_{0.71}$ & $61.21_{0.56}$ & $68.00_{0.34}$ \\
CAP \cite{wang2025handling}
    & $76.80_{0.84}$ & $89.96_{0.46}$ & $89.53_{0.70}$
    & $81.48_{0.45}$ & $83.32_{0.58}$ & $88.82_{0.18}$
    & $55.29_{0.31}$ & $62.33_{0.58}$ & $69.55_{0.51}$ \\
\midrule
\textbf{Ours}
    & \textbf{77.60}$_{1.28}$ & \textbf{90.88}$_{0.30}$ & \textbf{90.69}$_{0.19}$
    & \textbf{82.62}$_{0.26}$ & \textbf{85.33}$_{0.79}$ & \textbf{88.92}$_{0.21}$
    & \textbf{58.28}$_{0.29}$ & \textbf{63.35}$_{0.83}$ & \textbf{69.92}$_{0.32}$ \\
\bottomrule
\end{tabular}
}

\vspace{0.3cm}

\resizebox{\textwidth}{!}{
\begin{tabular}{lccccccccc}
\toprule
 & \multicolumn{3}{c}{EuroSAT} & \multicolumn{3}{c}{CUB} & \multicolumn{3}{c}{FGVC-Aircraft} \\
\cmidrule(lr){2-4} \cmidrule(lr){5-7} \cmidrule(lr){8-10}
Methods & UL & SSL & TRZSL & UL & SSL & TRZSL & UL & SSL & TRZSL \\
\midrule
CLIP \cite{radford2021learning}
    & $32.88_{0.00}$ & $32.88_{0.00}$ & $30.54_{0.00}$
    & $51.82_{0.00}$ & $51.82_{0.00}$ & $51.57_{0.00}$
    & $17.58_{0.00}$ & $17.58_{0.00}$ & $17.86_{0.00}$ \\
FPL \cite{menghini2023enhancing}
    & $48.96_{1.49}$ & $62.05_{1.64}$ & $53.70_{26.87}$
    & $53.04_{0.53}$ & $55.29_{0.59}$ & $55.44_{0.20}$
    & $16.62_{0.67}$ & $20.02_{0.77}$ & $17.55_{0.37}$ \\
GRIP \cite{menghini2023enhancing}
    & $57.21_{1.77}$ & $58.66_{2.64}$ & $92.33_{0.69}$
    & $51.42_{0.21}$ & $56.65_{0.33}$ & $59.48_{0.38}$
    & $15.22_{0.71}$ & $16.98_{0.82}$ & $26.08_{0.25}$ \\
CPL \cite{zhang2024candidate}
    & $67.26_{0.47}$ & $77.51_{0.80}$ & $93.78_{0.12}$
    & $53.47_{0.36}$ & $58.53_{0.24}$ & $66.20_{0.50}$
    & $18.35_{0.27}$ & $22.48_{0.63}$ & $30.86_{0.70}$ \\
CAP \cite{wang2025handling}
    & $75.01_{1.94}$ & $92.78_{0.34}$ & $96.64_{0.27}$
    & $55.76_{0.26}$ & $58.04_{0.23}$ & $61.35_{0.05}$
    & $18.42_{0.13}$ & $21.79_{0.16}$ & $29.03_{0.49}$ \\
\midrule
\textbf{Ours}
    & \textbf{95.30}$_{0.26}$ & \textbf{94.70}$_{0.16}$ & \textbf{98.17}$_{0.34}$
    & \textbf{60.72}$_{0.50}$ & \textbf{66.12}$_{0.72}$ & \textbf{72.30}$_{0.22}$
    & \textbf{18.84}$_{1.92}$ & \textbf{25.21}$_{0.30}$ & \textbf{31.65}$_{1.33}$ \\
\bottomrule
\end{tabular}
}
\end{sc}
\end{small}
\end{center}
\end{table*}

\subsection{Main Results}
\label{sec:main_results}

Table~\ref{tab:main_results} summarizes the quantitative results on six benchmarks under UL, SSL, and TRZSL. Across all 18 dataset-setting combinations, SAST improves over the strongest baseline. These results support the central claim of our paper: accounting for prediction instability improves pseudo-label-based CLIP adaptation, especially when text anchors are misaligned with target-domain visual structure.

\textbf{Unsupervised Learning (UL).}
SAST shows substantial improvements in the fully unlabeled setting, where adaptation depends entirely on pseudo-label supervision and prediction instability is particularly costly because mistakes can be repeatedly recycled. With no labeled anchor to correct a drifting boundary, the improvements on EuroSAT, Flowers102, and CUB are consistent with the intended role of SWIP, namely substituting temporal stability for supervision when no labeled anchor is available.

\textbf{Semi-Supervised Learning (SSL).}
SAST also improves consistently in the low-label setting. The gains are largest on CUB and FGVC-Aircraft, two fine-grained datasets where two labeled samples per class are insufficient to resolve target-domain ambiguity. This suggests that stability-aware prototype refresh is most useful when scarce supervision cannot reliably anchor the decision boundary by itself.

\textbf{Transductive Zero-Shot Learning (TRZSL).}
SAST remains effective in generalized transductive zero-shot learning. The improvements are largest on CUB, EuroSAT, and FGVC-Aircraft, where the seen/unseen split introduces stronger semantic mismatch than in the UL and SSL settings. This suggests that stability-aware anchoring helps maintain the seen/unseen balance under this mismatch.
\subsection{Analysis}
\label{sec:analysis}

We now analyze why SAST works, focusing on component contributions, the relation between stability and confidence, and sensitivity to hyperparameters and backbone scale. Additional analyses, including the CTPA robustness analysis and the overhead, are provided in Appendices~\ref{llm_experiments} and~\ref{app:complexity}.

\begin{figure*}[t]
    \centering
    \begin{subfigure}[b]{0.32\textwidth}
        \centering
        \includegraphics[width=\linewidth]{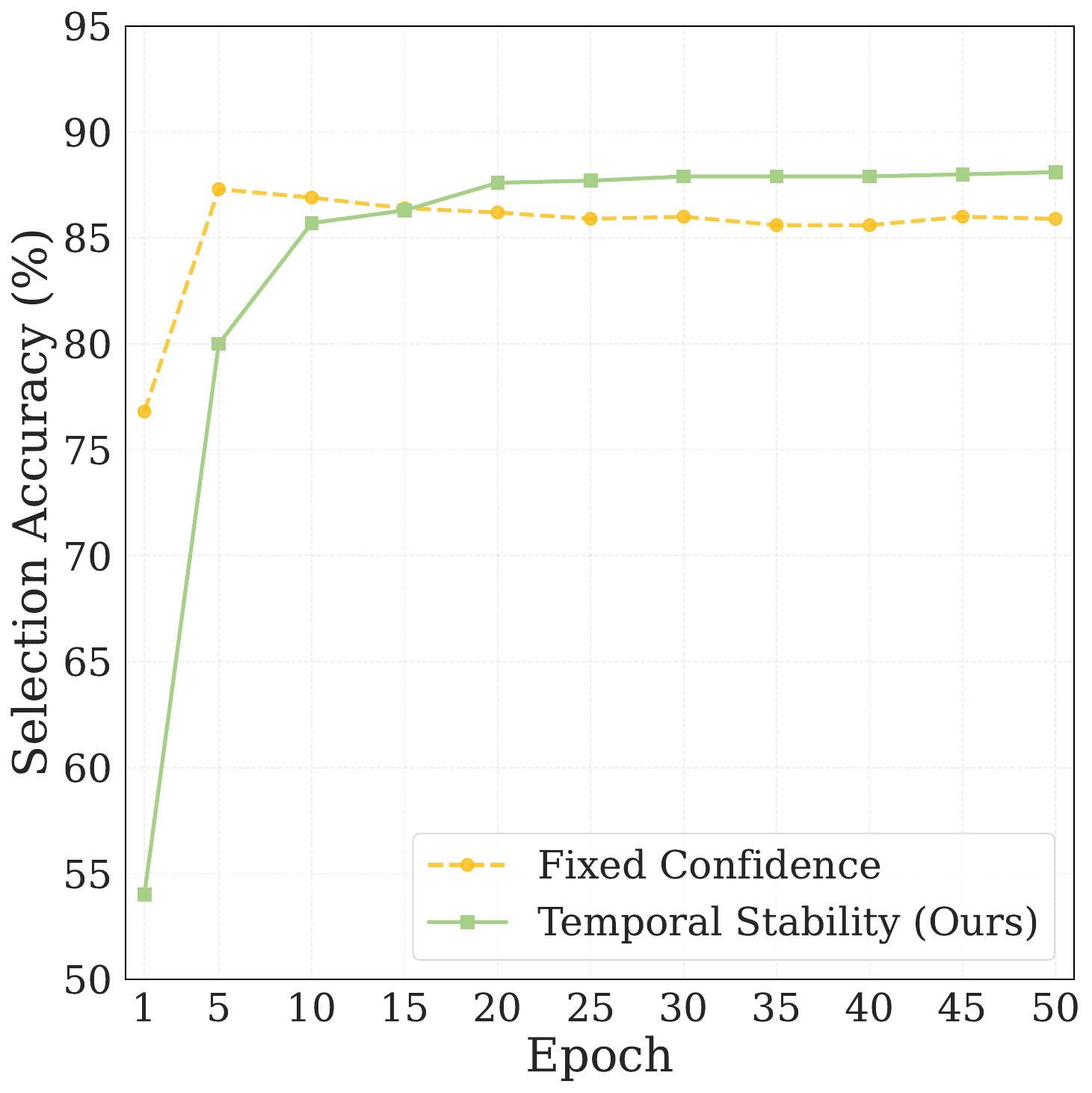}
        \caption{Prototype selections.}
        \label{fig:stability_analysis}
    \end{subfigure}
    \begin{subfigure}[b]{0.32\textwidth}
        \centering
        \includegraphics[width=\linewidth]{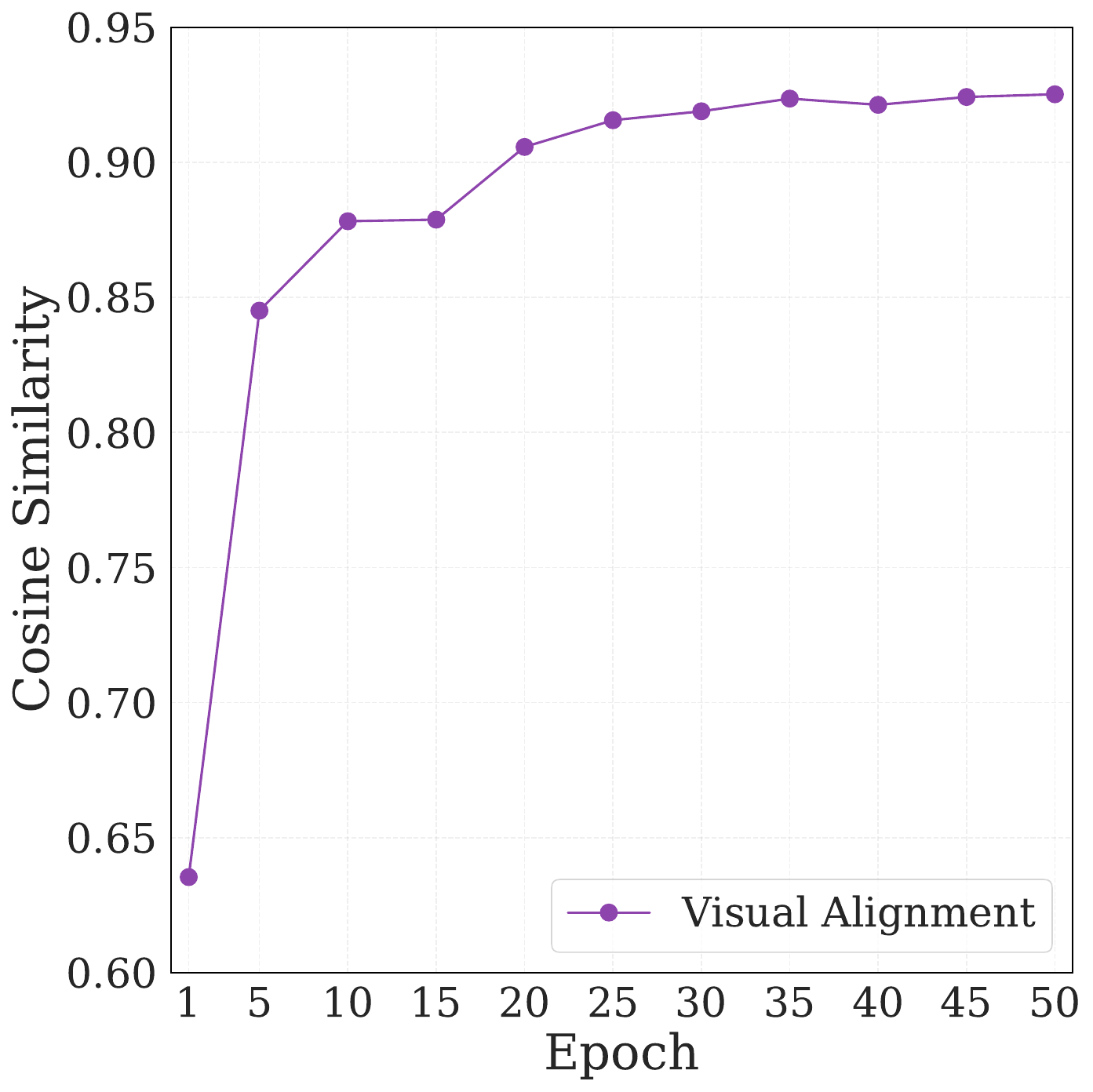}
        \caption{Prototype alignment.}
        \label{fig:visual_align}
    \end{subfigure}
    \begin{subfigure}[b]{0.32\textwidth}
        \centering
        \includegraphics[width=\linewidth]{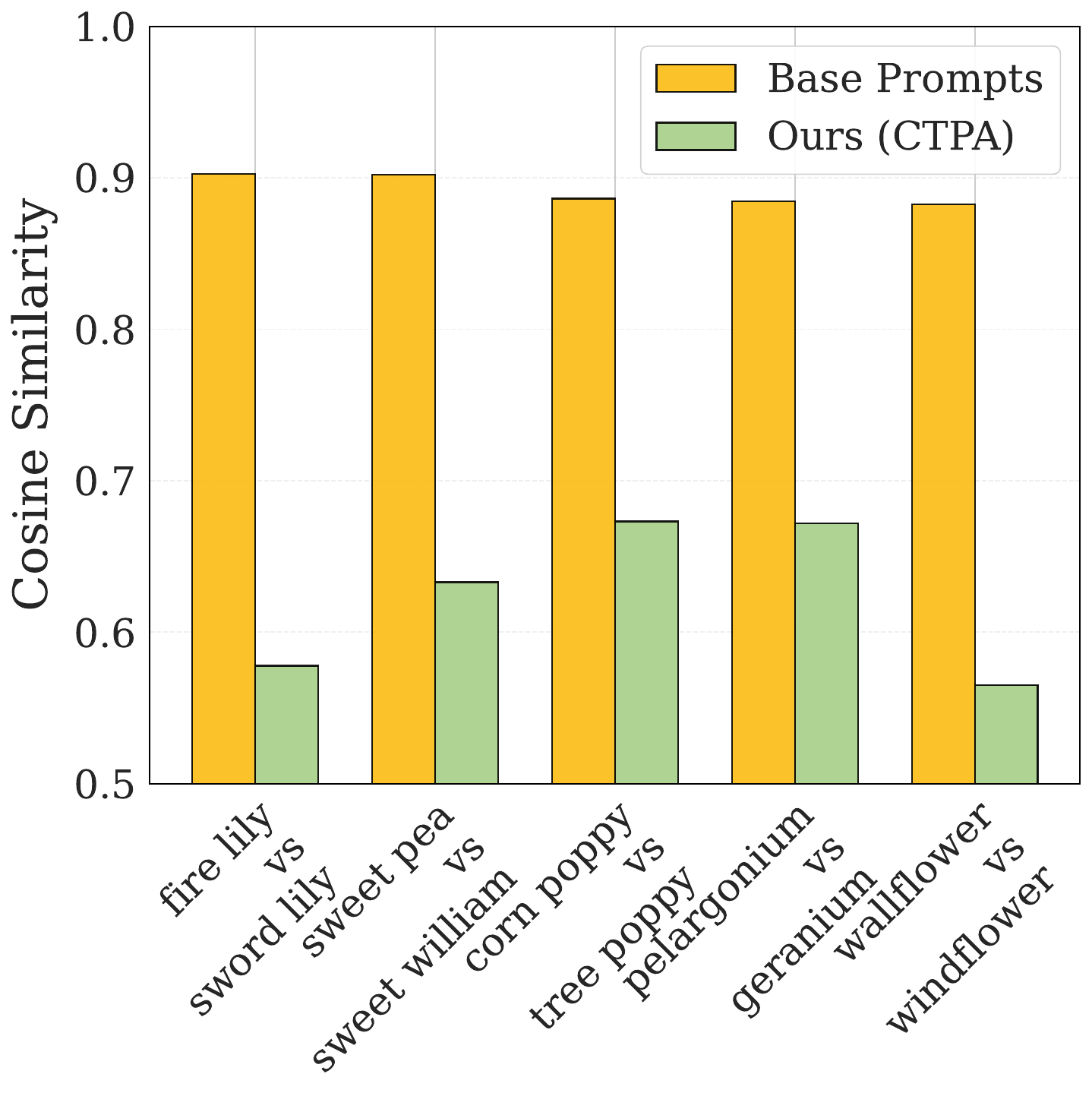}
        \caption{Similarity reduction.}
        \label{fig:text_sim}
    \end{subfigure}
    \caption{Analysis on the effects of the components of SAST.}
    \label{fig:ana_main}
\end{figure*}

\textbf{Is Temporal Stability an Effective Selection Signal?}
To evaluate whether temporal stability provides a reliable criterion for constructing image prototypes in SWIP, we track pseudo-label accuracy during training on RESISC45 under the UL setting. The pseudo-label accuracy is computed against ground-truth labels for analysis only and is not used during training. We compare the standard confidence-based strategy with our stability-based ranking in Figure~\ref{fig:stability_analysis}. The confidence baseline quickly plateaus at around 86\%, while the stability-based strategy shows a stronger upward trend after the early stage of training and eventually exceeds 88\% pseudo-label accuracy. This result supports our claim that temporal stability is an effective criterion for identifying reliable target samples.

\textbf{Do SWIP Prototypes Better Align with Target Visual Clusters?}
To examine whether SWIP progressively captures the target-domain visual structure, we visualize the alignment between the learned image prototypes and the target visual clusters on RESISC45 under the SSL setting. Figure~\ref{fig:visual_align} reports the average cosine similarity between the learned prototypes and the class-wise target cluster centers, where the latter are computed using ground-truth labels. The initial similarity (0.64) reflects the alignment under the pre-trained CLIP before adaptation, indicating a substantial mismatch between the initial CLIP representation and the target-domain feature distribution. As training proceeds, the similarity rises steadily and stabilizes around 0.93. This trend suggests that SWIP progressively selects more stable candidates and builds prototypes that better match the target visual structure, which is consistent with its role in mitigating prediction instability under anchoring mismatch.

\textbf{Does CTPA Improve Text-Side Stability Geometry?}
We analyze whether CTPA reduces confusion-induced instability by enlarging local text-side separation among semantically nearby classes on Flowers102. Figure~\ref{fig:text_sim} shows the cosine similarity of the text features for five representative confusing class pairs before and after prompt augmentation. In these cases, the enhanced prompts yield lower similarity than the base prompts. Notable reductions are observed for pairs such as \textit{fire lily} vs.\ \textit{sword lily} ($\Delta=-0.325$) and \textit{wallflower} vs.\ \textit{windflower} ($\Delta=-0.318$). These results indicate that confusion-aware attribute augmentation pushes nearby text features farther apart, improving local text-side separation and thereby helping reduce prediction switches and improving temporal stability.

\begin{table}[t]
\centering

\begin{minipage}[t]{0.48\linewidth}
\centering
\caption{Ablation of SWIP and CTPA.}
\label{tab:ablation_components}
\begin{small}
\begin{sc}
\begin{tabular}{cc|ccc}
\toprule
SWIP & CTPA & EuroSAT & CUB & DTD \\
\midrule
$\times$ & $\times$ & 91.7 & 58.4 & 59.6 \\
$\times$ & \checkmark & 91.8 & 59.9 & 62.8 \\
\checkmark & $\times$ & 93.2 & 64.7 & 61.1 \\
\checkmark & \checkmark & \textbf{94.7} & \textbf{66.1} & \textbf{63.4} \\
\bottomrule
\end{tabular}
\end{sc}
\end{small}
\end{minipage}
\hfill
\begin{minipage}[t]{0.48\linewidth}
\centering
\caption{Swap experiment for text prompt.}
\label{tab:ctpa_swap}
\begin{small}
\begin{sc}
\begin{tabular}{lcc}
\toprule
Method & CUB & RESISC45 \\
\midrule
CAP & 58.0 & 83.3 \\
SAST + CAP prompt & 64.0 & 86.2 \\
CAP + our CTPA & 56.9 & 83.7 \\
SAST & 66.1 & 85.3 \\
\bottomrule
\end{tabular}
\end{sc}
\end{small}
\end{minipage}

\end{table}

\begin{table}[t]
\centering
\caption{CTPA activation statistics and its accuracy gains.}
\label{tab:ctpa_activation}
\begin{small}
\begin{sc}
\begin{tabular}{lccc}
\toprule
Dataset & Fraction $\mathcal{N}(c)\neq\emptyset$ & Mean $|\mathcal{N}(c)|$ & CTPA gain \\
\midrule
Flowers102 & 0.40 & 1.0 & $+0.5$ \\
CUB & 0.70 & 3.7 & $+2.1$ \\
EuroSAT & 1.00 & 6.6 & $+2.7$ \\
RESISC45 & 0.97 & 18.1 & $-0.9$ \\
DTD & 0.98 & 24.8 & $+4.3$ \\

\bottomrule
\end{tabular}
\end{sc}
\end{small}
\end{table}

\begin{table}[t]
\centering

\begin{minipage}[t]{0.48\linewidth}
\centering
\caption{Single-branch vs. fused prediction.}
\label{tab:single_vs_fused}
\begin{small}
\begin{sc}
\begin{tabular}{lccc}
\toprule
Dataset & Text & Visual & Fused \\
\midrule
CUB            & 56.9 & 58.0 & \textbf{60.7} \\
RESISC45       & 81.3 & 81.9 & \textbf{82.6} \\
DTD            & 57.9 & 55.4 & \textbf{58.3} \\
Flowers102     & 75.7 & 77.1 & \textbf{77.6} \\
\bottomrule
\end{tabular}
\end{sc}
\end{small}
\end{minipage}
\hfill
\begin{minipage}[t]{0.48\linewidth}
\centering
\caption{Comparison of fusion strategies.}
\label{tab:ablation_fusion}
\begin{small}
\begin{sc}
\begin{tabular}{lcc}
\toprule
Fusion variant & DTD & CUB \\
\midrule
SWIP:CTPA = 1:0 & 60.1 & 64.7 \\
SWIP:CTPA = 0:1 & 62.9 & 59.8 \\
SWIP:CTPA = 0.5:0.5 & 62.4 & 64.1 \\
Global learned scalar $\alpha$ & 58.2 & 63.9 \\
Ours & \textbf{63.4} & \textbf{66.1} \\
\bottomrule
\end{tabular}
\end{sc}
\end{small}
\end{minipage}
\end{table}
\textbf{Component Ablation.}
To examine the roles of the components in SAST, we perform ablations on EuroSAT, CUB, and DTD under the SSL setting. Table~\ref{tab:ablation_components} reports the results of removing SWIP and CTPA from the full framework. Both components improve over the baseline, and their combination yields the best performance, supporting their complementary roles in the proposed stability-aware framework. This complementarity is not uniform, and we observe dataset-dependent trends: SWIP contributes more on CUB and EuroSAT, where stability-based visual anchoring is particularly helpful under fine-grained ambiguity, while CTPA brings larger gains on DTD, where reducing text-side confusion appears more beneficial. The ablation indicates that SWIP accounts for most of the gain on ambiguity-sensitive datasets such as CUB, supporting our claim that stability-aware anchoring is the primary driver of SAST. CTPA provides complementary refinement for confusion-prone classes.

\textbf{Attribution of CTPA Gains.}
The ablation shows that CTPA contributes complementarily to SWIP, but it does not rule out that the gain comes from stronger text descriptions rather than from the stability-aware integration. We therefore cross-combine the two adaptation frameworks (CAP\citep{wang2025handling} and SAST) with the two text prompt sources (the CAP-style prompt and our CTPA attributes), keeping all other components unchanged. As shown in Table~\ref{tab:ctpa_swap}, SAST consistently outperforms CAP under both prompt sources, whereas feeding our attributes directly to CAP yields no comparable gain. This indicates that the improvement stems primarily from how the text branch is integrated into the stability-aware framework, rather than from any prompt text alone. We therefore view CTPA as a semantic refinement whose benefit is realized through the stability-guided adaptation framework.

\textbf{When Does CTPA Help?}
To understand when CTPA is beneficial, we compute two statistics: the fraction of classes with a non-empty confusion set $\mathcal{N}(c)$, which reflects how often CTPA is activated, and the mean $|\mathcal{N}(c)|$, which characterizes the degree of semantic ambiguity, both under $\delta=0.8$. As shown in Table~\ref{tab:ctpa_activation}, with the exception of RESISC45, CTPA brings larger gains as more classes participate in refinement and as the neighborhoods become denser, with DTD and EuroSAT showing the clearest gains. For RESISC45, we attribute this deviation to the compositional nature of its categories, such as \textit{railway} vs.\ \textit{railway station}, which share substantial semantic components rather than distinctive object attributes. Even when prompted for discriminative attributes, the candidate pool retains considerable semantic overlap. Since CTPA performs confusion-aware selection rather than attribute generation, its benefit is bounded by the discriminability of the candidate attribute pool.

\textbf{Single-Branch vs. Fused Prediction.}
Table~\ref{tab:single_vs_fused} compares the single-branch and fused predictors on four datasets under the UL setting. Across all datasets, the adaptive fusion consistently exceeds the better single branch. This suggests that the two streams are not redundant and that the adaptive fusion can effectively exploit their combination. For example, on CUB, the fused predictor reaches 60.7\%, outperforming both the text branch (56.9\%) and the visual branch (58.0\%). A similar trend is observed on RESISC45, DTD, and Flowers102, indicating that the benefit of fusion is consistent.

\textbf{Comparison of Fusion Strategies.}
Having established that fusing the two branches is beneficial, we next ask how they should be fused. 
We compare the proposed adaptive fusion with several alternative integration strategies on DTD and CUB under the SSL setting. 
As shown in Table~\ref{tab:ablation_fusion}, our method consistently outperforms two single-stream variants (SWIP-only and CTPA-only), arithmetic average, and globally learned fusion rule. 
The global learned scalar yields inferior performance, suggesting that a single dataset-level trade-off is insufficient to capture the instance-dependent reliability of the two streams. These results support our design of estimating branch reliability on a per-sample basis.

\begin{table}[t]
\centering

\begin{minipage}[t]{0.5\linewidth}
\centering
\caption{Replacing stability with confidence.}
\label{tab:stability_vs_confidence}
\begin{small}
\begin{sc}
\begin{tabular}{lccc}
\toprule
Config & RESISC45 & DTD & CUB \\
\midrule
Stab. in both & \textbf{83.6} & \textbf{57.5} & \textbf{60.7} \\
SWIP $\to$ conf. & 76.6 & 55.6 & 57.9 \\
Fusion $\to$ conf. & 81.5 & 56.9 & 60.4 \\
Conf. in both & 75.6 & 55.7 & 57.7 \\
\bottomrule
\end{tabular}
\end{sc}
\end{small}
\end{minipage}
\hfill
\begin{minipage}[t]{0.48\linewidth}
\centering
\caption{Correlations.}
\label{tab:proxy_corr}
\begin{small}
\begin{sc}
\begin{tabular}{lcc}
\toprule
Proxy & Acc. & Stab. \\
\midrule
Stab. & 0.566 & 1.000 \\
Conf. & 0.518 & 0.624 \\
Entropy & -0.513 & -0.606 \\
Margin & 0.485 & 0.767 \\
\bottomrule
\end{tabular}
\end{sc}
\end{small}
\end{minipage}
\end{table}

\textbf{Is Stability More Than Confidence?}
A natural question is whether temporal stability merely reparameterizes confidence. We examine this question at two levels: the component level and the sample level. At the component level, we replace stability with confidence in the two components that use it: SWIP prototype construction and adaptive fusion. 
As shown in Table~\ref{tab:stability_vs_confidence}, replacing either component reduces accuracy, and using confidence in both yields the lowest accuracy on all three datasets. 
When all components rely on confidence, the same signal is used for pseudo-label filtering, prototype selection, and fusion, making these components partially redundant and thereby reducing accuracy. This indicates that stability contributes information that confidence alone does not provide.

\begin{wrapfigure}[12]{r}{0.45\textwidth}
  \centering
  \vspace{-1\baselineskip}
  \includegraphics[width=\linewidth]{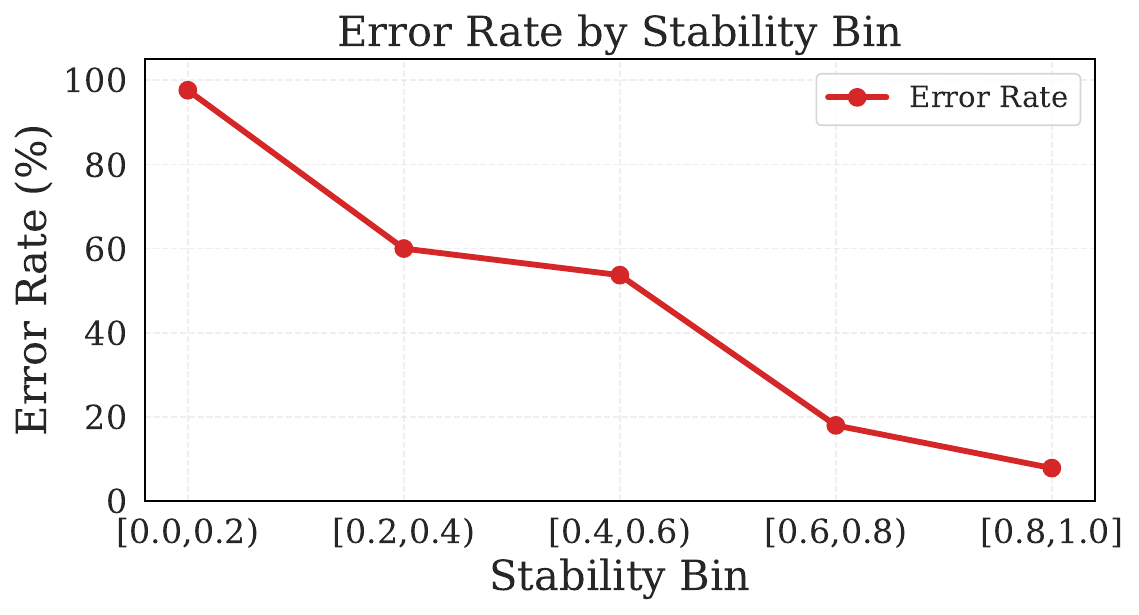}
  \caption{Pseudo-label error rate vs. stability.}
  \label{fig:error_rate}
\end{wrapfigure}

At the sample level, we analyze pseudo-label correctness on CUB by grouping target samples according to their final-round stability scores. 
As shown in Figure~\ref{fig:error_rate}, the pseudo-label error rate decreases monotonically as stability increases, from 97.6\% in the lowest-stability bin to 7.8\% in the highest-stability bin. 
We further condition on both signals. 
For high-confidence samples ($\text{confidence} \ge 0.9$), the overall accuracy is 84.2\% ($n=2021$). 
However, among low-confidence yet high-stability samples ($\text{confidence} < 0.9, S(\mathbf{x}) \ge 0.8$), accuracy remains remarkably high at 82.6\% ($n=310$), nearly comparable to that of high-confidence samples. 
This pattern indicates that stability identifies a subset of low-confidence predictions that are nonetheless reliable, a distinction that confidence cannot make.

\textbf{How Should Reliability Signals Be Assigned?} The above results show that stability captures information beyond confidence. Overall, these results suggest that stability is not a replacement for confidence-based filtering, nor merely a proxy for confidence, but a complementary structural signal: confidence serves as the per-sample filtering criterion, while stability determines which samples form visual anchors and how the two streams are weighted. Since high-stability samples are relatively limited in number, they are naturally suited to serve as anchors, where precision matters more than coverage, whereas high-confidence samples are much more abundant and can provide broader pseudo-label supervision. Table~\ref{tab:proxy_corr} quantifies this complementarity. Temporal stability shows the strongest correlation with pseudo-label accuracy (0.566) and is therefore used during training when prediction histories are available. Among the instantaneous proxies, confidence and entropy behave similarly (0.518 and 0.513) and are strongly anti-correlated with each other, indicating that they capture closely related uncertainty information. The logit margin is a notable case: although its correlation with accuracy (0.485) is slightly weaker than that of confidence, it shows the strongest correlation with temporal stability (0.767) among the instantaneous proxies. We attribute this to the fact that margin is closely tied to local decision-boundary robustness and hence to the difficulty of prediction switching. However, as a single-step signal, margin is more easily affected by samples that are consistently but incorrectly predicted, whereas temporal stability benefits from long-horizon aggregation and is therefore more robust as a reliability measure. When prediction histories are unavailable, as is the case for unseen test samples, we therefore retain confidence as the default single-pass proxy, while using temporal stability as the primary reliability signal during training.

\begin{table}[t]
\centering
\caption{Ablation of hyperparameters. Default values are shown in \textbf{bold}.}
\label{tab:hyperparam_ablation}
\begin{small}
\begin{sc}
\begin{tabular}{lccc}
\toprule
Parameter & Values & Flowers102 & FGVC-Aircraft \\
\midrule
$S_0$ & 0 / \textbf{0.5} / 1.0 & 90.3 / 90.9 / 90.6 & 24.6 / 25.2 / 24.9 \\
$\lambda_c$ & 0.05 / \textbf{0.1} / 0.2 & 90.9 / 90.9 / 90.6 & 24.9 / 25.2 / 25.1 \\
$k$ & 4 / \textbf{8} / 12 & 91.0 / 90.9 / 90.6 & 24.8 / 25.2 / 24.9 \\
$\delta$
& 0.7 / \textbf{0.8} / 0.9
& 90.7 / 90.9 / 91.0
& 25.2 / 25.2 / 24.4 \\
$\mu_{\text{init}}$ (Flowers102)
& 0.85 / \textbf{0.90} / 0.95
& 89.8 / 90.9 / 90.4
& -- \\

$\mu_{\text{init}}$ (FGVC-Aircraft)
& 0.50 / \textbf{0.60} / 0.70
& --
& 24.9 / 25.2 / 25.1 \\
momentum & 0.8 / \textbf{0.9} / 0.99 & 91.0 / 90.9 / 89.8 & 24.0 / 25.2 / 24.3 \\

\bottomrule
\end{tabular}
\end{sc}
\end{small}
\end{table}

\begin{table}[t]
\centering
\caption{Comparison of accuracy using ViT-L/14 as backbone. The best results are in \textbf{bold}.}
\label{tab:encoder_comparison}
\begin{small}
\begin{sc}
\begin{tabular}{lccc}
\toprule
\multicolumn{4}{c}{DTD} \\
\midrule
Methods & SSL & UL & TRZSL \\
\midrule
Zero-shot CLIP & 52.45 & 52.45 & 51.61 \\
FPL & 60.61 & 52.99 & 60.77 \\
GRIP & 60.91 & 54.40 & 64.92 \\
CPL & 69.82 & 57.20 & 71.97 \\
CAP & 70.26 & 65.85 & 73.61 \\
\textbf{Ours} & \textbf{70.64} & \textbf{66.38} & \textbf{75.40} \\
\bottomrule
\end{tabular}
\hfill
\begin{tabular}{lccc}
\toprule
\multicolumn{4}{c}{RESISC45} \\
\midrule
Methods & SSL & UL & TRZSL \\
\midrule
Zero-shot CLIP & 62.67 & 62.67 & 62.13 \\
FPL & 79.01 & 70.85 & 77.69 \\
GRIP & 81.53 & 76.86 & 86.88 \\
CPL & 87.75 & 80.88 & 89.73 \\
CAP & 88.09 & 84.46 & 90.21 \\
\textbf{Ours} & \textbf{91.16} & \textbf{85.34} & \textbf{93.02} \\
\bottomrule
\end{tabular}
\end{sc}
\end{small}
\end{table}

\textbf{Hyperparameter Sensitivity.}
We evaluate the robustness of SAST to key hyperparameters, including the initial disagreement frequency $S_0$, the penalty weight $\lambda_c$, the initial prototype quota $k$, the text-similarity threshold $\delta$ used to define confusion neighbors, the base pseudo-label threshold $\mu_{\text{init}}$, and the momentum. Each hyperparameter is varied independently while keeping the others fixed at their default values. As shown in Table~\ref{tab:hyperparam_ablation}, SAST remains reasonably stable across the tested ranges on Flowers102 and FGVC-Aircraft, suggesting that SAST does not rely heavily on precise tuning.

\textbf{Comparison Across Different Encoders.}
To further evaluate the generalizability of SAST across different backbone scales, we replace the default ViT-B/32 with a larger ViT-L/14 visual backbone and conduct additional experiments on DTD and RESISC45 under SSL, UL, and TRZSL settings. As shown in Table~\ref{tab:encoder_comparison}, our method consistently outperforms all baselines across all settings on both datasets. These results suggest that the proposed stability-aware design is not tied to a specific encoder scale, and can continue to provide gains even when starting from a stronger visual representation.

\section{Conclusion}

We study prediction instability in CLIP self-training through cross-modal anchoring mismatch: classifying target-domain visual features against pre-trained text anchors misaligned with the local target structure yields a fragile decision geometry under cosine-based classification. We propose a Stability-Aware Self-Training (SAST) framework, built around Stability-Weighted Image Prototypes (SWIP) for constructing visual anchors from temporally stable samples, and Confusion-Aware Text Prompt Augmentation (CTPA) for improving text-side separation among confusing categories, combined by an adaptive fusion rule. Our experiments across six benchmarks and three learning settings show that explicitly modeling prediction instability consistently improves CLIP adaptation.
\par}

\begin{ack}
This work was supported by the National Natural Science Foundation of China (62676119, 62236003), Guangdong Provincial Key Laboratory of Artificial General Intelligence and Large Models (2026B1212010005), Guangdong Science and Technology Program (2025B0101130003), Shenzhen College Stability Support Plan (GXWD20231130151329002), and Shenzhen Science and Technology Program (KQTD20240729102207002, KQTD2025120109171700).
\end{ack}

\bibliographystyle{plainnat}
\bibliography{neurips_2026}

@inproceedings{guo2017calibration,
  title={On calibration of modern neural networks},
  author={Guo, Chuan and Pleiss, Geoff and Sun, Yu and Weinberger, Kilian Q},
  booktitle={International Conference on Machine Learning},
  pages={1321--1330},
  year={2017},
  organization={PMLR}
}

@inproceedings{zhang2024candidate,
  title={Candidate Pseudolabel Learning: Enhancing Vision-Language Models by Prompt Tuning with Unlabeled Data},
  author={Zhang, Jiahan and Wei, Qi and Liu, Feng and Feng, Lei},
  booktitle={International Conference on Machine Learning},
  pages={60004--60020},
  year={2024},
  organization={PMLR}
}

@article{menghini2023enhancing,
  title={Enhancing clip with clip: Exploring pseudolabeling for limited-label prompt tuning},
  author={Menghini, Cristina and Delworth, Andrew and Bach, Stephen},
  journal={Advances in Neural Information Processing Systems},
  volume={36},
  pages={60984--61007},
  year={2023}
}

@inproceedings{wang2025handling,
  title={Handling Imbalanced Pseudolabels for Vision-Language Models with Concept Alignment and Confusion-Aware Calibrated Margin},
  author={Wang, Yuchen and Bai, Xuefeng and Li, Xiucheng and Guan, Weili and Nie, Liqiang and Chen, Xinyang},
  booktitle={International Conference on Machine Learning},
  pages={62357--62372},
  year={2025},
  organization={PMLR}
}

@inproceedings{radford2021learning,
  title={Learning transferable visual models from natural language supervision},
  author={Radford, Alec and Kim, Jong Wook and Hallacy, Chris and Ramesh, Aditya and Goh, Gabriel and Agarwal, Sandhini and Sastry, Girish and Askell, Amanda and Mishkin, Pamela and Clark, Jack and others},
  booktitle={International Conference on Machine Learning},
  pages={8748--8763},
  year={2021},
  organization={PmLR}
}

@article{li2021align,
  title={Align before fuse: Vision and language representation learning with momentum distillation},
  author={Li, Junnan and Selvaraju, Ramprasaath and Gotmare, Akhilesh and Joty, Shafiq and Xiong, Caiming and Hoi, Steven Chu Hong},
  journal={Advances in Neural Information Processing Systems},
  volume={34},
  pages={9694--9705},
  year={2021}
}

@article{zhou2022learning,
  title={Learning to prompt for vision-language models},
  author={Zhou, Kaiyang and Yang, Jingkang and Loy, Chen Change and Liu, Ziwei},
  journal={International Journal of Computer Vision},
  volume={130},
  number={9},
  pages={2337--2348},
  year={2022},
  publisher={Springer}
}

@article{cai2025attribute,
  title={Attribute-based visual reprogramming for vision-language models},
  author={Cai, Chengyi and Ye, Zesheng and Feng, Lei and Qi, Jianzhong and Liu, Feng},
  journal={arXiv preprint arXiv:2501.13982},
  year={2025}
}

@inproceedings{zhang2022tip,
  title={Tip-adapter: Training-free adaption of clip for few-shot classification},
  author={Zhang, Renrui and Zhang, Wei and Fang, Rongyao and Gao, Peng and Li, Kunchang and Dai, Jifeng and Qiao, Yu and Li, Hongsheng},
  booktitle={European Conference on Computer Vision},
  pages={493--510},
  year={2022},
  organization={Springer}
}

@article{huang2022unsupervised,
  title={Unsupervised prompt learning for vision-language models},
  author={Huang, Tony and Chu, Jack and Wei, Fangyun},
  journal={arXiv preprint arXiv:2204.03649},
  year={2022}
}

@inproceedings{zhou2022conditional,
  title={Conditional prompt learning for vision-language models},
  author={Zhou, Kaiyang and Yang, Jingkang and Loy, Chen Change and Liu, Ziwei},
  booktitle={Proceedings of the IEEE/CVF Conference on Computer Vision and Pattern Recognition},
  pages={16816--16825},
  year={2022}
}

@inproceedings{lee2023read,
  title={Read-only prompt optimization for vision-language few-shot learning},
  author={Lee, Dongjun and Song, Seokwon and Suh, Jihee and Choi, Joonmyeong and Lee, Sanghyeok and Kim, Hyunwoo J},
  booktitle={Proceedings of the IEEE/CVF International Conference on Computer Vision},
  pages={1401--1411},
  year={2023}
}

@article{laine2016temporal,
  title={Temporal ensembling for semi-supervised learning},
  author={Laine, Samuli and Aila, Timo},
  journal={arXiv preprint arXiv:1610.02242},
  year={2016}
}

@article{chen2023softmatch,
  title={Softmatch: Addressing the quantity-quality trade-off in semi-supervised learning},
  author={Chen, Hao and Tao, Ran and Fan, Yue and Wang, Yidong and Wang, Jindong and Schiele, Bernt and Xie, Xing and Raj, Bhiksha and Savvides, Marios},
  journal={arXiv preprint arXiv:2301.10921},
  year={2023}
}

@article{sohn2020fixmatch,
  title={Fixmatch: Simplifying semi-supervised learning with consistency and confidence},
  author={Sohn, Kihyuk and Berthelot, David and Carlini, Nicholas and Zhang, Zizhao and Zhang, Han and Raffel, Colin A and Cubuk, Ekin Dogus and Kurakin, Alexey and Li, Chun-Liang},
  journal={Advances in Neural Information Processing Systems},
  volume={33},
  pages={596--608},
  year={2020}
}

@inproceedings{khattak2023maple,
  title={Maple: Multi-modal prompt learning},
  author={Khattak, Muhammad Uzair and Rasheed, Hanoona and Maaz, Muhammad and Khan, Salman and Khan, Fahad Shahbaz},
  booktitle={Proceedings of the IEEE/CVF Conference on Computer Vision and Pattern Recognition},
  pages={19113--19122},
  year={2023}
}

@inproceedings{jia2022visual,
  title={Visual prompt tuning},
  author={Jia, Menglin and Tang, Luming and Chen, Bor-Chun and Cardie, Claire and Belongie, Serge and Hariharan, Bharath and Lim, Ser-Nam},
  booktitle={European Conference on Computer Vision},
  pages={709--727},
  year={2022},
  organization={Springer}
}

@inproceedings{nilsback2008automated,
  title={Automated flower classification over a large number of classes},
  author={Nilsback, Maria-Elena and Zisserman, Andrew},
  booktitle={2008 Sixth Indian Conference on Computer Vision, Graphics \& Image Processing},
  pages={722--729},
  year={2008},
  organization={IEEE}
}

@article{wah2011caltech,
  title={The caltech-ucsd birds-200-2011 dataset},
  author={Wah, Catherine and Branson, Steve and Welinder, Peter and Perona, Pietro and Belongie, Serge},
  year={2011},
  publisher={California Institute of Technology}
}

@article{cheng2017remote,
  title={Remote sensing image scene classification: Benchmark and state of the art},
  author={Cheng, Gong and Han, Junwei and Lu, Xiaoqiang},
  journal={Proceedings of the IEEE},
  volume={105},
  number={10},
  pages={1865--1883},
  year={2017},
  publisher={IEEE}
}

@inproceedings{cimpoi2014describing,
  title={Describing textures in the wild},
  author={Cimpoi, Mircea and Maji, Subhransu and Kokkinos, Iasonas and Mohamed, Sammy and Vedaldi, Andrea},
  booktitle={Proceedings of the IEEE Conference on Computer Vision and Pattern Recognition},
  pages={3606--3613},
  year={2014}
}

@article{helber2019eurosat,
  title={Eurosat: A novel dataset and deep learning benchmark for land use and land cover classification},
  author={Helber, Patrick and Bischke, Benjamin and Dengel, Andreas and Borth, Damian},
  journal={IEEE Journal of Selected Topics in Applied Earth Observations and Remote Sensing},
  volume={12},
  number={7},
  pages={2217--2226},
  year={2019},
  publisher={IEEE}
}

@article{maji2013fine,
  title={Fine-grained visual classification of aircraft},
  author={Maji, Subhransu and Rahtu, Esa and Kannala, Juho and Blaschko, Matthew and Vedaldi, Andrea},
  journal={arXiv preprint arXiv:1306.5151},
  year={2013}
}

@article{bahng2022exploring,
  title={Exploring visual prompts for adapting large-scale models},
  author={Bahng, Hyojin and Jahanian, Ali and Sankaranarayanan, Swami and Isola, Phillip},
  journal={arXiv preprint arXiv:2203.17274},
  year={2022}
}

@article{zhang2021flexmatch,
  title={Flexmatch: Boosting semi-supervised learning with curriculum pseudo labeling},
  author={Zhang, Bowen and Wang, Yidong and Hou, Wenxin and Wu, Hao and Wang, Jindong and Okumura, Manabu and Shinozaki, Takahiro},
  journal={Advances in Neural Information Processing Systems},
  volume={34},
  pages={18408--18419},
  year={2021}
}

@inproceedings{cascante2021curriculum,
  title={Curriculum labeling: Revisiting pseudo-labeling for semi-supervised learning},
  author={Cascante-Bonilla, Paola and Tan, Fuwen and Qi, Yanjun and Ordonez, Vicente},
  booktitle={Proceedings of the AAAI Conference on Artificial Intelligence},
  volume={35},
  number={8},
  pages={6912--6920},
  year={2021}
}

@inproceedings{zhai2022lit,
  title={Lit: Zero-shot transfer with locked-image text tuning},
  author={Zhai, Xiaohua and Wang, Xiao and Mustafa, Basil and Steiner, Andreas and Keysers, Daniel and Kolesnikov, Alexander and Beyer, Lucas},
  booktitle={Proceedings of the IEEE/CVF Conference on Computer Vision and Pattern Recognition},
  pages={18123--18133},
  year={2022}
}

@article{wang2022freematch,
  title={Freematch: Self-adaptive thresholding for semi-supervised learning},
  author={Wang, Yidong and Chen, Hao and Heng, Qiang and Hou, Wenxin and Fan, Yue and Wu, Zhen and Wang, Jindong and Savvides, Marios and Shinozaki, Takahiro and Raj, Bhiksha and others},
  journal={arXiv preprint arXiv:2205.07246},
  year={2022}
}

@article{gao2024clip,
  title={Clip-adapter: Better vision-language models with feature adapters},
  author={Gao, Peng and Geng, Shijie and Zhang, Renrui and Ma, Teli and Fang, Rongyao and Zhang, Yongfeng and Li, Hongsheng and Qiao, Yu},
  journal={International Journal of Computer Vision},
  volume={132},
  number={2},
  pages={581--595},
  year={2024},
  publisher={Springer}
}

@article{liang2022mind,
  title={Mind the gap: Understanding the modality gap in multi-modal contrastive representation learning},
  author={Liang, Victor Weixin and Zhang, Yuhui and Kwon, Yongchan and Yeung, Serena and Zou, James Y},
  journal={Advances in Neural Information Processing Systems},
  volume={35},
  pages={17612--17625},
  year={2022}
}

@article{schrodi2024two,
  title={Two Effects, One Trigger: On the Modality Gap, Object Bias, and Information Imbalance in Contrastive Vision-Language Models},
  author={Schrodi, Simon and Hoffmann, David T and Argus, Max and Fischer, Volker and Brox, Thomas},
  journal={arXiv preprint arXiv:2404.07983},
  year={2024}
}

@article{he2024gradual,
  title={Gradual domain adaptation: Theory and algorithms},
  author={He, Yifei and Wang, Haoxiang and Li, Bo and Zhao, Han},
  journal={Journal of Machine Learning Research},
  volume={25},
  number={361},
  pages={1--40},
  year={2024}
}

@article{shao2026capt,
  title={CAPT: Confusion-Aware Prompt Tuning for Reducing Vision-Language Misalignment},
  author={Shao, Maoyuan and Gao, Yutong and Huang, Xinyang and Zhu, Chuang and Sun, Lijuan and Nan, Guoshun},
  journal={arXiv preprint arXiv:2603.02557},
  year={2026}
}

@inproceedings{tian2024argue,
  title={Argue: Attribute-guided prompt tuning for vision-language models},
  author={Tian, Xinyu and Zou, Shu and Yang, Zhaoyuan and Zhang, Jing},
  booktitle={Proceedings of the IEEE/CVF Conference on Computer Vision and Pattern Recognition},
  pages={28578--28587},
  year={2024}
}

@article{long2024training,
  title={Training-free unsupervised prompt for vision-language models},
  author={Long, Sifan and Wang, Linbin and Zhao, Zhen and Tan, Zichang and Wu, Yiming and Wang, Shengsheng and Wang, Jingdong},
  journal={arXiv preprint arXiv:2404.16339},
  year={2024}
}

\clearpage
\appendix

\section*{Appendix Contents}
\begin{itemize}
    \item \hyperref[app:instability_analysis]{A. More Analysis on Prediction Instability}
    \item \hyperref[app:interpretation]{B. A Geometric Interpretation of Temporal Prediction Switch}
    \item \hyperref[app:ctpa_llm]{C. LLM Usage in CTPA}
    \item \hyperref[app:settings]{D. Training Settings}
    \item \hyperref[app:complexity]{E. Complexity, Overhead, and Statistical Significance}
    \item \hyperref[app:limitation]{F. Limitation}
    \item \hyperref[app:social_impact]{G. Social Impact}
\end{itemize}

\section{More Analysis on Prediction Instability}
\label{app:instability_analysis}

\subsection{CLIP Self-Training versus a Unimodal Classifier}
\label{app:clip_vs_unimodal}

To assess whether the observed instability is specific to cross-modal adaptation rather than a generic property of self-training, we compare CLIP self-training with a standard unimodal classifier under matched settings. The unimodal classifier uses a ResNet-50 backbone pre-trained on ImageNet with a linear classification head, trained on the same target-domain data with the same rule and budget. The only intended difference is the prediction mechanism: a single visual encoder with a learned linear head, versus a frozen vision-language model with text anchors.

We report the temporal switch rate, i.e., the average fraction of unlabeled target samples whose predicted class changes between consecutive epochs, computed over the same epoch range for both models. Under this matched protocol, CLIP self-training exhibits a switch rate of $0.205$, compared with $0.133$ for the unimodal classifier. This gap suggests that cross-modal anchoring introduces an additional source of instability beyond the noise typically associated with pseudo-label self-training.

\subsection{Measurement Protocol}
\label{app:diag_protocol}

Unless otherwise specified, all experiments in this section are conducted on Flowers102 under the SSL protocol, using the same labeled support set $\mathcal{D}_{\mathrm{L}}$, unlabeled set $\mathcal{D}_{\mathrm{UL}}$, CLIP ViT-B/32 backbone, and self-training framework. Both settings use the same SSL supervision budget, namely 2 shots per class.

\paragraph{Matched-support anchor diagnostics.}

We fix the target split and the support set $\mathcal{D}_{\mathrm{L}}$, and compare
(i) standard text anchors from CLIP text embeddings under the base prompt, and
(ii) target-aligned anchors constructed from $\mathcal{D}_{\mathrm{L}}$.

For the visual-anchor setting, we encode each support image (2-shot per class) using the pre-trained CLIP image encoder $\phi(\cdot)$ and compute a class-wise image prototype by averaging the normalized support features:
\begin{equation}
\mathbf{w}_c^{\mathrm{vis}}=
\mathrm{Normalize}\!\left(
\frac{1}{|\mathcal{D}_{L,c}|}
\sum_{\mathbf{x}\in \mathcal{D}_{L,c}} \phi(\mathbf{x})
\right),
\end{equation}
where $\mathcal{D}_{L,c}$ denotes the labeled support images of class $c$. These prototypes form a target-aligned visual classifier weight matrix
$\mathbf{W}^{\mathrm{vis}}=[\mathbf{w}_1^{\mathrm{vis}},\ldots,\mathbf{w}_C^{\mathrm{vis}}]^\top$.
Classification logits are computed by cosine similarity,
\begin{equation}
z_c^{\mathrm{vis}}(\mathbf{x})
=
\mathrm{Normalize}(\phi(\mathbf{x}))^\top
\mathrm{Normalize}(\mathbf{w}_c^{\mathrm{vis}}),
\end{equation}

\paragraph{Geometry metrics.}
We use two geometric metrics:
(i) angular margin, i.e., the difference between top-1 and top-2 cosine similarities; and
(ii) approximate feature switch distance, i.e., the smallest targeted perturbation norm in feature space that changes the predicted class. Larger values indicate a more robust local decision geometry.

The approximate feature switch distance is computed by an iterative targeted attack in feature space: starting from a clean feature, we optimize an additive perturbation toward the runner-up class and record the first $\ell_2$ norm at which the predicted class changes. If no switch occurs within the attack budget, the sample is marked as non-switched.

\begin{figure*}
    \centering
    \begin{subfigure}[b]{0.32\textwidth}
        \centering
        \includegraphics[width=\linewidth]{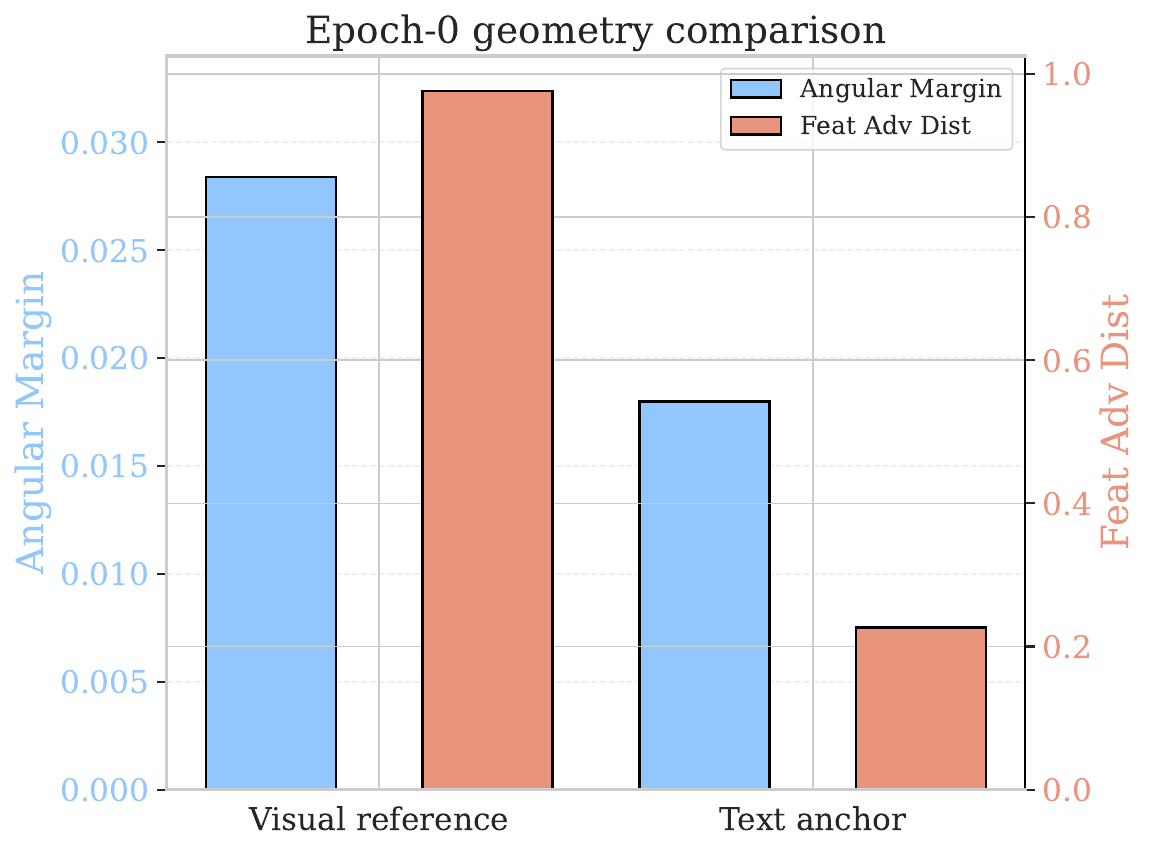}
        \caption{Epoch-0 geometry.}
        \label{fig:instability_morea}
    \end{subfigure}
    \begin{subfigure}[b]{0.32\textwidth}
        \centering
        \includegraphics[width=\linewidth]{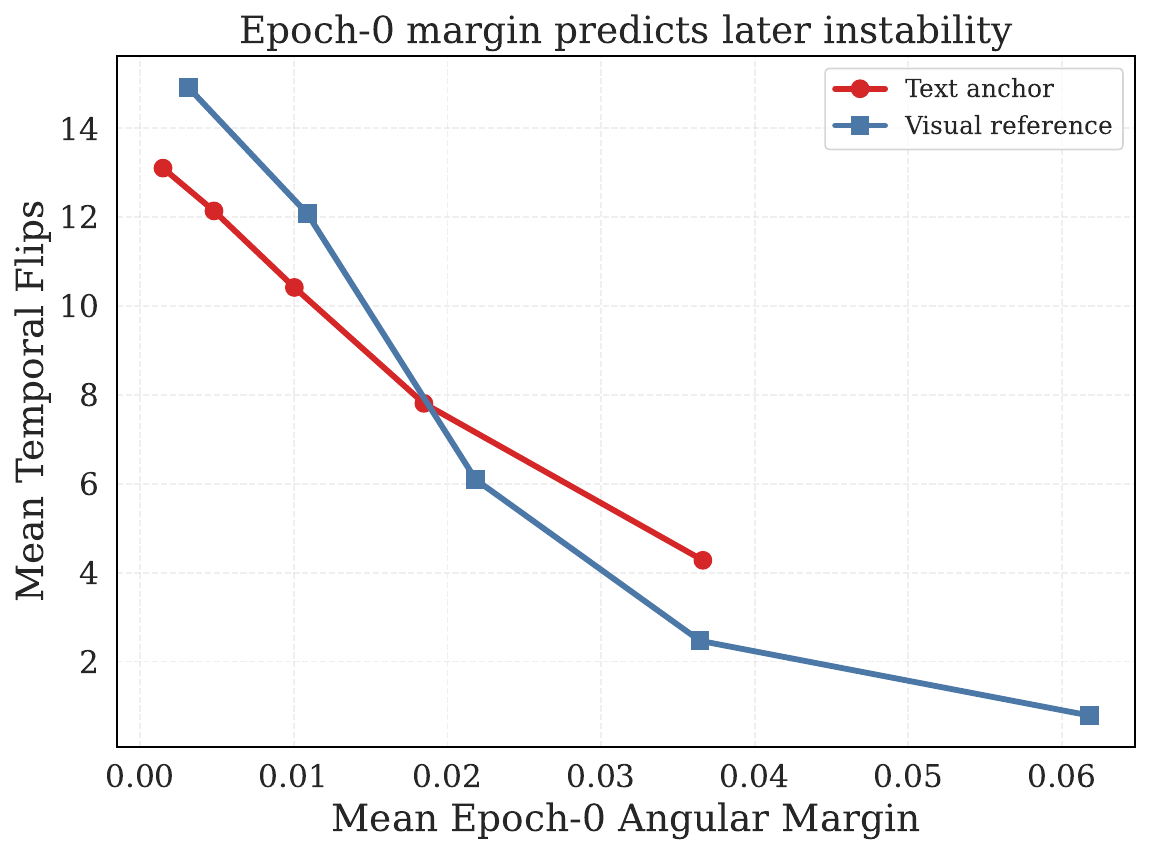}
        \caption{Initial margin vs.\ later switches.}
        \label{fig:instability_moreb}
    \end{subfigure}
    \begin{subfigure}[b]{0.32\textwidth}
        \centering
        \includegraphics[width=\linewidth]{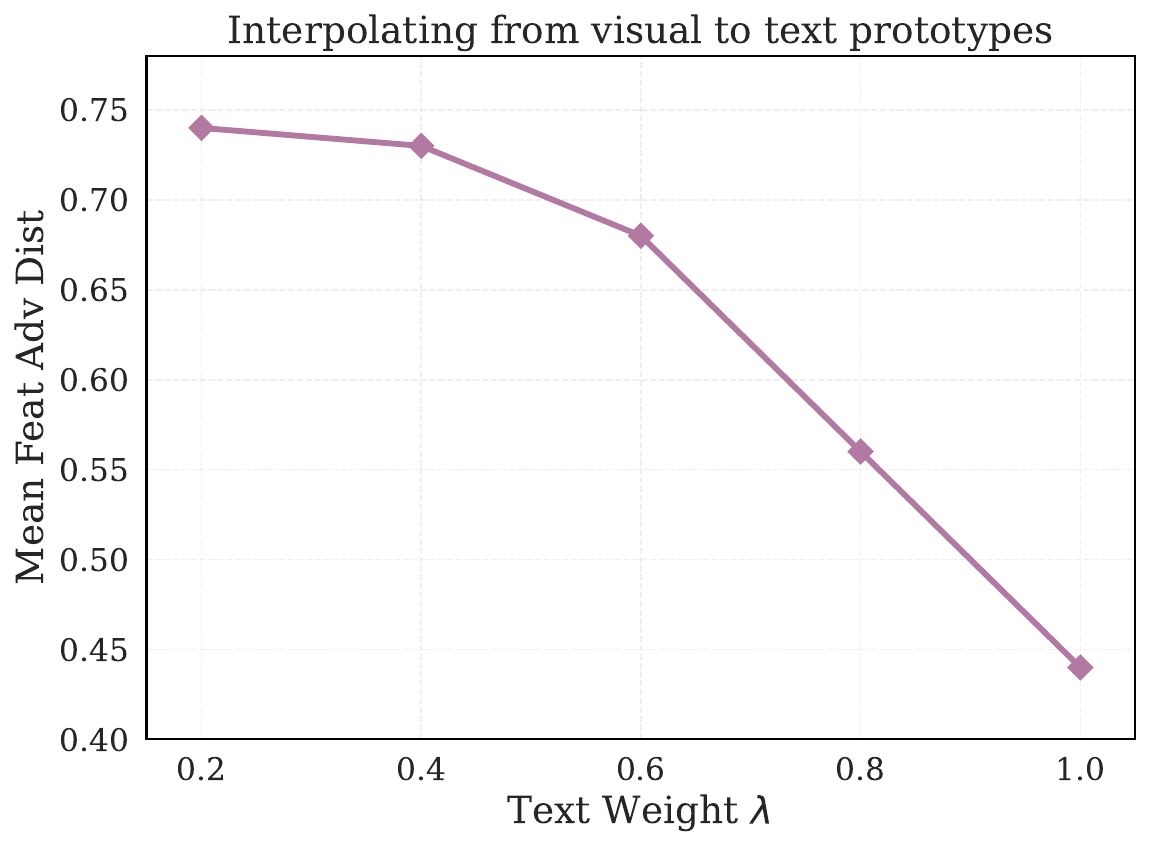}
        \caption{Robustness under interpolation.}
        \label{fig:instability_morec}
    \end{subfigure}
    \caption{Diagnostic analyses of prediction instability.}
    \label{fig:instability_more}
\end{figure*}

\begin{table}[t]
\centering
\caption{Matched diagnostic training summary on Flowers102.}
\label{tab:diag_traj}
\begin{small}
\begin{sc}
\begin{tabular}{lcc}
\toprule
Metric & Text anchor & Visual reference \\
\midrule
Epoch-0 acc. & 60.64 & 77.07 \\
Final acc. & 84.78 & 87.69 \\
Epoch-0 margin & 0.0180 & 0.0284 \\
Final margin & 0.0360 & 0.0593 \\
Epoch-0 switch dist. & 0.2263 & 0.9762 \\
Final switch dist. & 0.3346 & 1.0736 \\
Final switch rate & 15.49\% & 7.94\% \\
\bottomrule
\end{tabular}
\end{sc}
\end{small}
\end{table}

\subsection{Initialization Geometry and Training Stability}
\label{app:matched_dynamics}

We compare text anchors and target-aligned visual anchors under matched settings, so that only anchor construction differs.

At epoch~0, before self-training starts, the visual reference has a mean angular
margin about $1.58\times$ larger, an approximate feature switch distance more than
$4\times$ larger, and an epoch-0 accuracy about $16$ points higher than the text
anchor baseline (Figures~\ref{fig:instability_morea} and~\ref{fig:dy2}). This
suggests that the source of class anchors strongly affects the initial local
decision geometry on target-domain samples.

The visual reference remains consistently more stable throughout adaptation,
ending with a lower switch rate ($7.94\%$ vs.\ $15.49\%$) and higher accuracy
(87.69\% vs.\ 84.78\%) (Table~\ref{tab:diag_traj}). Target-aligned anchoring
thus provides more stable pseudo-label trajectories than standard text anchoring.

We further group target samples into five bins by their epoch-0 angular margin and measure the mean number of temporal switches accumulated during training.
Figure~\ref{fig:instability_moreb} shows a clear monotonic trend for both initializations: samples with larger epoch-0 margins undergo fewer switches later
in training. Under text anchors, the mean number of switches decreases from 13.10
in the lowest-margin bin to 4.28 in the highest; under the visual reference, it
drops from 14.92 to 0.79. The same trend holds when using approximate feature
switch distance instead of angular margin. These results suggest that early local
geometry is a useful indicator of which samples are likely to remain reliable
during self-training.

\subsection{Interpolating between Target-Aligned and Text Anchors}
\label{app:interp}

The previous analyses compare two endpoint references: purely visual and purely text anchors. To examine whether stability changes gradually with anchor construction, we interpolate between the two. For each class $c$, we form
\begin{equation}
\mathbf{w}_c^{(\alpha)}=
\mathrm{Normalize}\!\big((1-\alpha)\mathbf{w}_c^{\mathrm{vis}}+\alpha \mathbf{w}_c^{\mathrm{text}}\big),
\end{equation}
where $\alpha=0$ corresponds to fully visual anchors and $\alpha=1$ to purely text anchors.

As shown in Figure~\ref{fig:instability_morec}, the mean feature switch distance decreases monotonically as the classifier becomes more text-dominated, from about 0.74 at $\alpha=0$ to 0.44 at $\alpha=1$. This graded dependence is consistent with anchor construction: as class references become less target-aligned and more text-dominated, local robustness on target-domain samples weakens progressively.

\subsection{CTPA and Local Text Geometry}
\label{app:geometry_flip}

The analyses above explain why visual anchoring can improve stability. We next complement this view from the text side and examine whether improving local text geometry also reduces instability. To avoid selection bias, we define a fixed confusion-prone subset once using the base-prompt text similarities before any attribute augmentation, and keep it unchanged when comparing prompt variants.

Before augmentation, the average switch frequency of samples associated with these confusing classes is 17.6\% higher than the overall dataset average; after applying CTPA, this relative excess decreases to 7.26\%. On this fixed subset in Flowers102, CTPA reduces the average pairwise similarity of confusing class pairs (0.8478 → 0.7387) and increases the corresponding local margin (0.1633 → 0.2613), accompanied by a reduction in temporal switch rate from 10.7\% to 9.1\% (about a 15\% relative decrease).

These results support the intended role of CTPA: improving local text-side separation precisely in those neighborhoods where prediction switch is most concentrated. Taken together, the analyses in this appendix support a coherent picture: anchoring mismatch is associated with fragile local decision geometry and higher prediction instability in CLIP self-training. Improving visual anchoring through stability-guided prototypes is associated with lower switch and more reliable pseudo-labels, while text-side refinement provides a complementary effect that is most beneficial in confusion-dense label spaces and is realized in combination with visual adaptation.

\section{A Geometric Interpretation of Temporal Prediction Switch}
\label{app:interpretation}

This section gives a mechanism-level account of why target-domain visual features can exhibit fragile local decision geometry when classified against fixed text anchors. Our argument links class-conditional anchor offset to (i) a smaller effective angular margin, (ii) a smaller local logit gap, (iii) higher sensitivity to epoch-to-epoch perturbations, and thus (iv) a higher tendency of prediction switch. This is consistent with the matched-support diagnostics in Appendix~\ref{app:instability_analysis}.

\subsection{CLIP Classification on the Unit Hypersphere}
\label{app:definitions}

Let $\tilde{\boldsymbol{\phi}}(\mathbf{x})$ and $\tilde{\boldsymbol{\psi}}_c$ denote the $\ell_2$-normalized image and class-$c$ text features. CLIP computes the class logit by scaled cosine similarity,
\begin{equation}
    z_c(\mathbf{x}) = \frac{\tilde{\boldsymbol{\phi}}(\mathbf{x})^\top \tilde{\boldsymbol{\psi}}_c}{\tau}
    = \frac{\cos \theta_c}{\tau},
\end{equation}
where $\tau$ is the temperature and $\theta_c$ is the angle between the two features. For a competing pair of classes $(c,c')$, define the pairwise margin
\begin{equation}
    m_{c,c'}(\mathbf{x}) = \langle\tilde{\boldsymbol{\phi}}(\mathbf{x}), \tilde{\boldsymbol{\psi}}_c - \tilde{\boldsymbol{\psi}}_{c'}\rangle.
\end{equation}
Since the logits are cosine scores scaled by $\tau$, the corresponding pairwise logit gap is exactly
\begin{equation}
    z_c(\mathbf{x}) - z_{c'}(\mathbf{x}) = m_{c,c'}(\mathbf{x})/\tau.
    \label{eq:logit_gap_margin}
\end{equation}

\subsection{Class-Conditional Anchor Offset}
\label{app:center_offset}

Inspired by prior work on modality gap~\citep{liang2022mind}, we use a class-conditional anchor offset between target visual features and text anchors as a geometric measure for anchoring mismatch. Our focus is classification-time geometry on a target domain, where each sample is evaluated against fixed text anchors. For each class $c$, we define
\begin{equation}
    \Delta_{\mathrm{anchor}}^{(c)}
    =
    \left\|
    \bar{\boldsymbol{\phi}}_c - \tilde{\boldsymbol{\psi}}_c
    \right\|_2,
    \qquad
    \bar{\boldsymbol{\phi}}_c
    =
    \mathbb{E}_{\mathbf{x}\sim \mathcal{D}_c}
    \left[
    \tilde{\boldsymbol{\phi}}(\mathbf{x})
    \right],
\end{equation}
where $\mathcal{D}_c$ denotes the target-domain samples of class $c$. This quantity measures how far, on average, the target visual cluster of class $c$ is displaced from its text anchor. When $\Delta_{\mathrm{anchor}}^{(c)}$ is large, the text anchor becomes less dominant relative to nearby competing classes, making the local class ranking more fragile on target-domain samples.

\subsection{Anchor Offset and Pairwise Margin}
\label{app:center_to_margin}

We next relate the class-conditional anchor offset to the expected pairwise margin under the text classifier. Taking expectation of the pairwise margin over $\mathbf{x}\sim\mathcal{D}_c$ and writing the class center as $\bar{\boldsymbol{\phi}}_c = \tilde{\boldsymbol{\psi}}_c + \mathbf{e}_c$ gives
\begin{align}
    \mathbb{E}_{\mathbf{x}\sim\mathcal{D}_c}
    \big[
    m_{c,c'}(\mathbf{x})
    \big]
    &=
    1 - \left\langle \tilde{\boldsymbol{\psi}}_c, \tilde{\boldsymbol{\psi}}_{c'} \right\rangle
    +
    \left\langle
    \mathbf{e}_c,
    \tilde{\boldsymbol{\psi}}_c - \tilde{\boldsymbol{\psi}}_{c'}
    \right\rangle.
    \label{eq:expected_margin_decomposition}
\end{align}
The first term depends only on the separation between the two text features and becomes small when $c$ and $c'$ are semantically similar. The second term captures how the target visual center is displaced relative to the text anchor: when $\mathbf{e}_c$ has a component toward nearby competing classes, this correction is negative and reduces the expected margin.

Applying Cauchy--Schwarz yields the lower bound
\begin{equation}
    \mathbb{E}_{\mathbf{x}\sim\mathcal{D}_c}
    \big[
    m_{c,c'}(\mathbf{x})
    \big]
    \ge
    1 - \left\langle \tilde{\boldsymbol{\psi}}_c, \tilde{\boldsymbol{\psi}}_{c'} \right\rangle
    -
    \|\mathbf{e}_c\|_2
    \cdot
    \|\tilde{\boldsymbol{\psi}}_c - \tilde{\boldsymbol{\psi}}_{c'}\|_2.
    \label{eq:margin_lower_bound}
\end{equation}
Since $\|\mathbf{e}_c\|_2 = \Delta_{\mathrm{anchor}}^{(c)}$, a larger class-conditional anchor offset weakens this lower bound, especially when the competing class $c'$ is already close in text space.

\subsection{From Margin to Logit Gap}
\label{app:margin_to_gap}

By Eq.~\eqref{eq:logit_gap_margin}, any reduction in pairwise margin directly reduces the corresponding logit gap. The small-angle case further explains why semantically or geometrically close class pairs often produce especially small gaps.

Let $\theta_c$ and $\theta_{c'}$ denote the angles from the image feature to the matched class and a competing class, respectively, and let $\Delta\theta = \theta_{c'} - \theta_c$. For small $\Delta\theta$,
\begin{equation}
    z_c - z_{c'}
    \approx
    \frac{\sin\theta_c \cdot \Delta\theta}{\tau},
    \label{eq:small_angle_gap}
\end{equation}
which follows from the first-order expansion of $\cos\theta_{c'} = \cos(\theta_c + \Delta\theta)$. This local form makes the small-gap regime explicit: when the matched class and a confusing competitor are angularly close, the corresponding logit gap under cosine classification is also small.

\paragraph{Empirical support for the small-angle regime.}
We performed an additional analysis on Flowers102 with zero-shot CLIP. For each sample whose local class ranking switches in our perturbation analysis in Appendix~\ref{app:instability_analysis}, we extract the top-1 and top-2 logits and compute the corresponding angles $\theta_c$ and $\theta_{c'}$. We find that the mean angular gap $\Delta\theta = \theta_{c'}-\theta_c$ of these samples is only $0.21^\circ$. Moreover, for Eq.~\eqref{eq:small_angle_gap}, the absolute residual of the first-order approximation is $2.00\times 10^{-6}$, accounting for only $0.06\%$ of the corresponding logit value. This suggests that the small-$\Delta\theta$ approximation is well aligned with the local regime most relevant to prediction switch, and thus serves as a sanity check for Eq.~\eqref{eq:small_angle_gap} in the switch-prone regime.

\subsection{From Small Logit Gap to Higher Local Sensitivity}
\label{app:gap_to_flip}

The following proposition formalizes a local robustness property of the pairwise logit gap: when the gap is small, only a small feature-space perturbation is needed to change the local pairwise ordering between a matched class and a nearby competitor. We consider an additive perturbation $\boldsymbol{\delta}$ in normalized feature space; re-normalization preserves the sign of the gap because the normalization factor is strictly positive, so the condition below also applies to the exact re-normalized cosine classifier.

\begin{proposition}
\label{prop:gap_flip}
Let
\begin{equation}
    g_{c,c'}(\mathbf{x}; \boldsymbol{\delta})
    =
    \frac{
    \left\langle
    \tilde{\boldsymbol{\phi}}(\mathbf{x}) + \boldsymbol{\delta},
    \tilde{\boldsymbol{\psi}}_c - \tilde{\boldsymbol{\psi}}_{c'}
    \right\rangle
    }{\tau}
\end{equation}
denote the pairwise logit gap between classes $c$ and $c'$ under perturbation $\boldsymbol{\delta}$, where $\tilde{\boldsymbol{\phi}}(\mathbf{x})$ and $\tilde{\boldsymbol{\psi}}_c$ are $\ell_2$-normalized image and text features. Then
\begin{equation}
    \left|
    g_{c,c'}(\mathbf{x}; \boldsymbol{\delta})
    -
    g_{c,c'}(\mathbf{x}; \mathbf{0})
    \right|
    \le
    \frac{
    \|\boldsymbol{\delta}\|_2
    \cdot
    \|\tilde{\boldsymbol{\psi}}_c - \tilde{\boldsymbol{\psi}}_{c'}\|_2
    }{\tau}.
    \label{eq:gap_lipschitz}
\end{equation}
In particular, if $g_{c,c'}(\mathbf{x}; \mathbf{0}) > 0$ and
\begin{equation}
    \|\boldsymbol{\delta}\|_2
    <
    \frac{
    \tau \, g_{c,c'}(\mathbf{x}; \mathbf{0})
    }{
    \|\tilde{\boldsymbol{\psi}}_c - \tilde{\boldsymbol{\psi}}_{c'}\|_2
    },
    \label{eq:flip_threshold_formal}
\end{equation}
then $g_{c,c'}(\mathbf{x}; \boldsymbol{\delta}) > 0$.
\end{proposition}

\begin{proof}
By definition,
\begin{equation}
    g_{c,c'}(\mathbf{x}; \boldsymbol{\delta})
    -
    g_{c,c'}(\mathbf{x}; \mathbf{0})
    =
    \frac{
    \left\langle
    \boldsymbol{\delta},
    \tilde{\boldsymbol{\psi}}_c - \tilde{\boldsymbol{\psi}}_{c'}
    \right\rangle
    }{\tau}.
\end{equation}
Cauchy--Schwarz gives Eq.~\eqref{eq:gap_lipschitz}. Now assume $g_{c,c'}(\mathbf{x}; \mathbf{0}) > 0$ and Eq.~\eqref{eq:flip_threshold_formal} holds. Then $|g(\boldsymbol{\delta}) - g(\mathbf{0})| < g(\mathbf{0})$, so
\begin{equation}
    g_{c,c'}(\mathbf{x}; \boldsymbol{\delta})
    \ge
    g_{c,c'}(\mathbf{x}; \mathbf{0})
    -
    \left|
    g_{c,c'}(\mathbf{x}; \boldsymbol{\delta})
    -
    g_{c,c'}(\mathbf{x}; \mathbf{0})
    \right|
    > 0.
\end{equation}
Hence the sign of the pairwise gap is preserved.
\end{proof}

Proposition~\ref{prop:gap_flip} gives a deterministic local robustness criterion: the critical perturbation radius in Eq.~\eqref{eq:flip_threshold_formal} is proportional to the original pairwise logit gap. Samples in a small-gap regime therefore admit a smaller robustness radius. In self-training, such perturbations need not be adversarial; they may arise from epoch-to-epoch feature drift, stochastic optimization, or feedback from imperfect pseudo-label supervision. Samples that repeatedly remain in a small-gap regime under text anchor classification are thus plausibly more prone to changing their top-1 class across epochs, exhibiting temporal prediction switch.

\subsection{Geometric Interpretation}
\label{app:geo_interpretation}

Text anchors partition the unit hypersphere into spherical Voronoi cells. A large class-conditional anchor offset does not mean that target samples sit near cell boundaries. It does mean that target features are, on average, less centered around their text anchors, so more of them can fall in off-center regions of the cell, where the margin against a nearby class is small. This matches the matched-support diagnostic in Appendix~\ref{app:instability_analysis}: swapping text anchors for visual anchors enlarges angular margin and feature-space robustness, and lowers switch rates during training.

Proposition~\ref{prop:gap_flip} shows that a small logit gap leaves little room before a perturbation flips the class ranking. The analysis above shows that anchor offset makes such small-gap configurations more common on target-domain samples. Anchoring mismatch therefore raises the prevalence of fragile configurations that are prone to temporal prediction switch, without determining the outcome for any single sample. This is the regime SAST is designed to detect and mitigate.

\subsection{Stability and Pseudo-Label Noise}
\label{app:stability_pl_noise}

The geometric analysis above explains why temporal prediction switch arises. Here we ask why reducing it helps adaptation. On CUB, the pseudo-label error rate falls monotonically with stability, from 97.6\% in the lowest-stability bin to 7.8\% in the highest (Figure~\ref{fig:error_rate}), and stability is complementary to confidence: among high-confidence samples, accuracy is 93.13\% for high-stability samples but only 70.55\% for low-stability ones. High-stability samples are thus a cleaner source of pseudo-label supervision than confidence alone would suggest.

Prioritizing them reduces the pseudo-label noise that enters self-training. Pseudo-label errors can accumulate across training steps and degrade target performance~\citep{he2024gradual}; lower switch rates indicate more stable predictions, which in turn yield more reliable pseudo-labels and less error accumulation.

\section{LLM Usage in CTPA}
\label{app:ctpa_llm}

\subsection{LLM Setup}
\label{app:ctpa_llm_setup}

We use \textbf{Gemini 2.5 Pro} as the LLM for offline attribute generation. For each class, the LLM returns a JSON list of candidate attributes, from which CTPA selects the subset that is least compatible with the confusion neighbors of the class according to Eq.~(6)--(7) in the main text. The LLM is instructed to produce attributes that emphasize what makes a class visually recognizable and different from other classes in the same domain.

In all settings, including TRZSL, the attribute pool is generated from class names only, without access to instance labels, ground-truth descriptions, or any held-out annotations. For unseen classes in TRZSL, attributes are generated before adaptation begins and use only the class names; no unseen images, pseudo-labels, or target-domain statistics are provided to the LLM. This matches the information available to the baselines. Generated attribute pools are released with the code.

\subsection{Prompt}
\label{app:ctpa_llm_prompt}

We use the following prompt template for offline attribute generation:

\begin{quote}
\small
Your goal is to create a single JSON object containing a list of 10--15 distinctive attributes for each class provided in the input. Distinctive attributes are features that uniquely identify a class and differentiate it from other classes within its domain. Focus on what makes it special, different, or easily recognizable compared to others.

Example:

Input: ``Wilson Warbler''

Output:
\{
  ``Wilson Warbler'': [
    ``The roundness of the cap on the top of the head'',
    ``The olive-green on the back is a slightly dark shade'',
    ``The little yellow bird with a black cap''
  ]
\}

New Task:

Now, generate the distinctive attributes for all the following classes. Ensure that you provide 10--15 attributes for each class in a single JSON object.
\end{quote}

\subsection{LLM- and attribute-level robustness of CTPA}
\label{llm_experiments}

\paragraph{Robustness to the Choice of LLM.} 
We repeat CTPA using attributes generated by Qwen3.5-0.8B, a 1B-scale model, under the same prompting protocol. Replacing Gemini-2.5-Pro with Qwen3.5-0.8B decreases average accuracy by only $0.8\%$ across the six datasets (Table~\ref{tab:llm_ablation}). The largest degradation appears on fine-grained datasets such as CUB and Flowers102, where the smaller model produces more generic attributes; on DTD and EuroSAT, performance remains close to Gemini-2.5-Pro. Even with this weaker LLM, CTPA still improves over CAP by $2.1\%$ on average, suggesting that the gain of SAST does not rely on a frontier LLM.

\begin{table}[t]
\centering
\caption{CTPA performance with different attribute-generating LLMs.}
\label{tab:llm_ablation}
\begin{small}
\begin{sc}
\begin{tabular}{lcccccc}
\toprule
LLM & CUB & Flowers102 & DTD & EuroSAT & RESISC45 & FGVC-Aircraft \\
\midrule
Gemini-2.5-Pro & 66.1 & 90.9 & 63.4 & 94.7 & 85.3 & 25.2 \\
Qwen3.5-0.8B & 64.5 & 90.0 & 62.5 & 93.9 & 85.1 & 24.8 \\
\bottomrule
\end{tabular}
\end{sc}
\end{small}
\end{table}

\paragraph{Which Types of Attributes Are Effective?} 
For a confusing class pair $(A,B)$, we measure the similarity drop
\begin{equation}
\Delta_{\mathrm{sim}} = \mathrm{sim}_{\mathrm{base}}(A,B) - \mathrm{sim}_{\mathrm{typed}}(A,B),
\end{equation}
when only one type of attribute is used. As shown in Table~\ref{tab:attribute_types}, the most effective type is domain- and pair-dependent: on Flowers102, color attributes produce the largest separation, while on CUB no single type dominates. Selecting a small discriminative subset also outperforms stacking all attributes, which supports the design of CTPA.

\begin{table}[t]
\centering
\caption{$\Delta_{\mathrm{sim}}$ by attribute types (10 confusing pairs each).}
\label{tab:attribute_types}
\begin{small}
\begin{sc}
\begin{tabular}{lcccccc}
\toprule
\multicolumn{7}{c}{Flowers102} \\
\midrule
Type & color & shape & texture & part & count & habit \\
\midrule
Mean $\Delta_{\mathrm{sim}}$ & 0.297 & 0.184 & 0.171 & 0.244 & 0.164 & 0.179 \\
\bottomrule
\end{tabular}

\vspace{0.2cm}

\begin{tabular}{lccccc}
\toprule
\multicolumn{6}{c}{CUB} \\
\midrule
Type & color & shape & texture & pattern & all attributes \\
\midrule
Mean $\Delta_{\mathrm{sim}}$ & 0.056 & 0.039 & 0.224 & 0.049 & 0.033 \\
\bottomrule
\end{tabular}
\end{sc}
\end{small}
\end{table}

\section{Training Settings}
\label{app:settings}

The detailed training settings used in the main experiments are summarized in Table~\ref{tab:exp_settings}.

\section{Complexity, Overhead, and Statistical Significance}
\label{app:complexity}

All measurements below are on DTD using a single NVIDIA A100 GPU under the same backbone and training budget.

\paragraph{Computational cost.}
SAST introduces two additional sources of overhead: (i) epoch-wise prototype refresh in SWIP, and (ii) dual-stream logit computation followed by fusion. The refresh is performed once per epoch with class-wise ranking, so its cost is moderate. Table~\ref{tab:complexity_main} reports the comparison: FLOPs are nearly identical to CAP ($188.297$ vs.\ $188.275$ GFLOPs), and training time is 64 min on DTD, comparable to CAP (62 min).

\paragraph{Memory and storage overhead.}
The additional memory cost comes from the prediction-history queue and the class-wise prototype matrix. The storage complexity is $\mathcal{O}(NQ + CD)$, where $Q \le 50$ is the history length and $C$ is the number of classes. Because the queue stores only predicted class labels and lightweight statistics rather than full features, increasing the budget beyond the current 8 GiB would not change what is stored or how stability is estimated; we therefore do not expect additional accuracy gains from a larger budget. Peak GPU memory is 8206 MiB, compared with 8019 MiB for CAP (Table~\ref{tab:complexity_main}).

\begin{table}[t]
\centering
\small
\setlength{\tabcolsep}{4pt}
\caption{Detailed settings.}
\label{tab:exp_settings}

\begin{tabular}{@{}l|c|c|c|c|c|c@{}}
    \toprule
     & Flowers102 & RESISC45 & DTD & CUB & EuroSAT & FGVC-Aircraft \\
    \midrule
    Class number & 102 & 45 & 47 & 200 & 10 & 100 \\
    Training set size & 2040 & 6300 & 3760 & 5594 & 27000 & 6667 \\
    Testing set size & 6149 & 25200 & 1880 & 5794 & 5000 & 3333 \\
    \midrule
    Network & \multicolumn{6}{c}{ViT-B/32} \\
    \hline
    Batch size & \multicolumn{6}{c}{32} \\
    \hline
    Epochs & \multicolumn{6}{c}{50} \\
    \hline
    Optimizer & \multicolumn{6}{c}{SGD} \\
    \hline
    SGD momentum & \multicolumn{6}{c}{0.9} \\
    \hline
    Learning rate & \multicolumn{6}{c}{0.02} \\
    \hline
    Weight decay & \multicolumn{6}{c}{0.05} \\
    \hline
    Scheduler & \multicolumn{6}{c}{CosineAnnealingLR} \\
    \bottomrule
\end{tabular}%

\end{table}

\begin{table}[t]
\centering

\begin{minipage}[t]{0.66\linewidth}
\centering
\caption{Complexity comparison.}
\label{tab:complexity_main}
\begin{small}
\begin{sc}
\begin{tabular}{lccc}
\toprule
Method & GFLOPs & Training time & Peak GPU memory \\
\midrule
CAP  & 188.275 & 62 min  & 8019 MiB \\
SAST & 188.297 & 64 min  & 8206 MiB \\
\bottomrule
\end{tabular}
\end{sc}
\end{small}
\end{minipage}
\hfill
\begin{minipage}[t]{0.32\linewidth}
\centering
\caption{Storage cost.}
\label{tab:complexity_buffer}
\begin{small}
\begin{sc}
\begin{tabular}{lc}
\toprule
Dataset & Cost \\
\midrule
CUB      & 4.5 MB \\
EuroSAT  & 21 MB \\
\bottomrule
\end{tabular}
\end{sc}
\end{small}
\end{minipage}

\end{table}

\paragraph{Statistical significance of small gains.}
For two settings where the improvement over CAP is small,
we repeat each run with 10 random seeds (1--10) under identical configurations.
Paired $t$-tests yield $p=0.0004$ (DTD-TRZSL, ViT-B/32, mean gain $+0.334$)
and $p=0.0276$ (DTD-SSL, ViT-L/14, mean gain $+0.457$), with Cohen's $d$
of 1.56 and 0.697, respectively. The absolute gains are small but statistically
significant and occur on already strong baselines.

\section{Limitation}
\label{app:limitation}

This work has several limitations. First, our analysis is partly formal at the local geometric level, but it is not yet a full theoretical characterization of self-training dynamics. In particular, the interaction among temporal stability, pseudo-label noise accumulation, branch-wise reliability, and representation drift remains to be studied more systematically. Second, although SAST performs well on the evaluated benchmarks, its scalability to larger label spaces, larger unlabeled pools, and broader deployment settings has not yet been fully established. In particular, settings where prediction histories are unavailable, costly to maintain, or only partially observable may require modified stability estimators. Third, because SAST is implemented as a PEFT-based adaptation method with pseudo-label self-training, its strongest gains should be expected in regimes where the target domain still lies within the transferable range of the pre-trained representation and the main bottleneck is adaptation rather than missing semantics. Accordingly, its ability to correct severe class-level mismatch inherited from pretraining may remain limited when the target domain is too far from CLIP pretraining or when the initial text-side semantics are strongly biased relative to the target domain. Exploring stronger combinations with explicit alignment or representation-level adaptation is an important direction for future work.

\section{Social Impact}
\label{app:social_impact}

By improving the reliability of self-training for vision-language models, this work may help reduce annotation cost and improve adaptation to new visual domains. However, better adaptation can also make such systems easier to deploy in high-stakes or weakly monitored settings. If the target data contain systematic biases, stability-aware filtering may preserve or even reinforce biased pseudo-labels rather than remove them. In addition, LLM-generated attributes may inherit biases from the language model. We therefore recommend careful evaluation before deployment, especially in sensitive or high-impact applications.
\newpage
\section*{NeurIPS Paper Checklist}

The checklist is designed to encourage best practices for responsible machine learning research, addressing issues of reproducibility, transparency, research ethics, and societal impact. Do not remove the checklist: {\bf The papers not including the checklist will be desk rejected.} The checklist should follow the references and follow the (optional) supplemental material.  The checklist does NOT count towards the page
limit. 

Please read the checklist guidelines carefully for information on how to answer these questions. For each question in the checklist:
\begin{itemize}
    \item You should answer \answerYes{}, \answerNo{}, or \answerNA{}.
    \item \answerNA{} means either that the question is Not Applicable for that particular paper or the relevant information is Not Available.
    \item Please provide a short (1--2 sentence) justification right after your answer (even for \answerNA). 
\end{itemize}

{\bf The checklist answers are an integral part of your paper submission.} They are visible to the reviewers, area chairs, senior area chairs, and ethics reviewers. You will also be asked to include it (after eventual revisions) with the final version of your paper, and its final version will be published with the paper.

The reviewers of your paper will be asked to use the checklist as one of the factors in their evaluation. While \answerYes{} is generally preferable to \answerNo{}, it is perfectly acceptable to answer \answerNo{} provided a proper justification is given (e.g., error bars are not reported because it would be too computationally expensive'' or ``we were unable to find the license for the dataset we used''). In general, answering \answerNo{} or \answerNA{} is not grounds for rejection. While the questions are phrased in a binary way, we acknowledge that the true answer is often more nuanced, so please just use your best judgment and write a justification to elaborate. All supporting evidence can appear either in the main paper or the supplemental material, provided in appendix. If you answer \answerYes{} to a question, in the justification please point to the section(s) where related material for the question can be found.

IMPORTANT, please:
\begin{itemize}
    \item {\bf Delete this instruction block, but keep the section heading ``NeurIPS Paper Checklist"},
    \item  {\bf Keep the checklist subsection headings, questions/answers and guidelines below.}
    \item {\bf Do not modify the questions and only use the provided macros for your answers}.
\end{itemize}


\begin{enumerate}

\item {\bf Claims}
    \item[] Question: Do the main claims made in the abstract and introduction accurately reflect the paper's contributions and scope?
    \item[] Answer: \answerYes{} 
    \item[] Justification: We include detailed information in Section~\ref{sec:intro}.
    \item[] Guidelines:
    \begin{itemize}
        \item The answer \answerNA{} means that the abstract and introduction do not include the claims made in the paper.
        \item The abstract and/or introduction should clearly state the claims made, including the contributions made in the paper and important assumptions and limitations. A \answerNo{} or \answerNA{} answer to this question will not be perceived well by the reviewers. 
        \item The claims made should match theoretical and experimental results, and reflect how much the results can be expected to generalize to other settings. 
        \item It is fine to include aspirational goals as motivation as long as it is clear that these goals are not attained by the paper. 
    \end{itemize}

\item {\bf Limitations}
    \item[] Question: Does the paper discuss the limitations of the work performed by the authors?
    \item[] Answer: \answerYes{}
    \item[] Justification: The paper includes a dedicated limitations section (Appendix~\ref{app:limitation}).
    \item[] Guidelines:
    \begin{itemize}
        \item The answer \answerNA{} means that the paper has no limitation while the answer \answerNo{} means that the paper has limitations, but those are not discussed in the paper. 
        \item The authors are encouraged to create a separate ``Limitations'' section in their paper.
        \item The paper should point out any strong assumptions and how robust the results are to violations of these assumptions (e.g., independence assumptions, noiseless settings, model well-specification, asymptotic approximations only holding locally). The authors should reflect on how these assumptions might be violated in practice and what the implications would be.
        \item The authors should reflect on the scope of the claims made, e.g., if the approach was only tested on a few datasets or with a few runs. In general, empirical results often depend on implicit assumptions, which should be articulated.
        \item The authors should reflect on the factors that influence the performance of the approach. For example, a facial recognition algorithm may perform poorly when image resolution is low or images are taken in low lighting. Or a speech-to-text system might not be used reliably to provide closed captions for online lectures because it fails to handle technical jargon.
        \item The authors should discuss the computational efficiency of the proposed algorithms and how they scale with dataset size.
        \item If applicable, the authors should discuss possible limitations of their approach to address problems of privacy and fairness.
        \item While the authors might fear that complete honesty about limitations might be used by reviewers as grounds for rejection, a worse outcome might be that reviewers discover limitations that aren't acknowledged in the paper. The authors should use their best judgment and recognize that individual actions in favor of transparency play an important role in developing norms that preserve the integrity of the community. Reviewers will be specifically instructed to not penalize honesty concerning limitations.
    \end{itemize}

\item {\bf Theory assumptions and proofs}
    \item[] Question: For each theoretical result, does the paper provide the full set of assumptions and a complete (and correct) proof?
    \item[] Answer: \answerYes{}
    \item[] Justification: The paper includes a local robustness proposition together with its assumptions and a complete proof in Appendix~\ref{app:gap_to_flip} (Proposition~\ref{prop:gap_flip}). The geometric interpretation and assumptions used for the analysis are described in Appendix~\ref{app:interpretation}.
    \item[] Guidelines:
    \begin{itemize}
        \item The answer \answerNA{} means that the paper does not include theoretical results. 
        \item All the theorems, formulas, and proofs in the paper should be numbered and cross-referenced.
        \item All assumptions should be clearly stated or referenced in the statement of any theorems.
        \item The proofs can either appear in the main paper or the supplemental material, but if they appear in the supplemental material, the authors are encouraged to provide a short proof sketch to provide intuition. 
        \item Inversely, any informal proof provided in the core of the paper should be complemented by formal proofs provided in appendix or supplemental material.
        \item Theorems and Lemmas that the proof relies upon should be properly referenced. 
    \end{itemize}

    \item {\bf Experimental result reproducibility}
    \item[] Question: Does the paper fully disclose all the information needed to reproduce the main experimental results of the paper to the extent that it affects the main claims and/or conclusions of the paper (regardless of whether the code and data are provided or not)?
    \item[] Answer: \answerYes{}
    \item[] Justification: The paper specifies the datasets, learning settings, backbone, training protocol, optimization setup, ablations, and diagnostic analyses needed to reproduce the reported results. Key implementation details are given in Section~\ref{sec:experiments} and Appendix~\ref{app:settings}.
    \item[] Guidelines:
    \begin{itemize}
        \item The answer \answerNA{} means that the paper does not include experiments.
        \item If the paper includes experiments, a \answerNo{} answer to this question will not be perceived well by the reviewers: Making the paper reproducible is important, regardless of whether the code and data are provided or not.
        \item If the contribution is a dataset and\slash or model, the authors should describe the steps taken to make their results reproducible or verifiable. 
        \item Depending on the contribution, reproducibility can be accomplished in various ways. For example, if the contribution is a novel architecture, describing the architecture fully might suffice, or if the contribution is a specific model and empirical evaluation, it may be necessary to either make it possible for others to replicate the model with the same dataset, or provide access to the model. In general. releasing code and data is often one good way to accomplish this, but reproducibility can also be provided via detailed instructions for how to replicate the results, access to a hosted model (e.g., in the case of a large language model), releasing of a model checkpoint, or other means that are appropriate to the research performed.
        \item While NeurIPS does not require releasing code, the conference does require all submissions to provide some reasonable avenue for reproducibility, which may depend on the nature of the contribution. For example
        \begin{enumerate}
            \item If the contribution is primarily a new algorithm, the paper should make it clear how to reproduce that algorithm.
            \item If the contribution is primarily a new model architecture, the paper should describe the architecture clearly and fully.
            \item If the contribution is a new model (e.g., a large language model), then there should either be a way to access this model for reproducing the results or a way to reproduce the model (e.g., with an open-source dataset or instructions for how to construct the dataset).
            \item We recognize that reproducibility may be tricky in some cases, in which case authors are welcome to describe the particular way they provide for reproducibility. In the case of closed-source models, it may be that access to the model is limited in some way (e.g., to registered users), but it should be possible for other researchers to have some path to reproducing or verifying the results.
        \end{enumerate}
    \end{itemize}

\item {\bf Open access to data and code}
    \item[] Question: Does the paper provide open access to the data and code, with sufficient instructions to faithfully reproduce the main experimental results, as described in supplemental material?
    \item[] Answer: \answerYes{}
    \item[] Justification: Our code is available at \url{https://github.com/lsy-2003/SAST}.
    \item[] Guidelines:
    \begin{itemize}
        \item The answer \answerNA{} means that paper does not include experiments requiring code.
        \item Please see the NeurIPS code and data submission guidelines (\url{https://neurips.cc/public/guides/CodeSubmissionPolicy}) for more details.
        \item While we encourage the release of code and data, we understand that this might not be possible, so \answerNo{} is an acceptable answer. Papers cannot be rejected simply for not including code, unless this is central to the contribution (e.g., for a new open-source benchmark).
        \item The instructions should contain the exact command and environment needed to run to reproduce the results. See the NeurIPS code and data submission guidelines (\url{https://neurips.cc/public/guides/CodeSubmissionPolicy}) for more details.
        \item The authors should provide instructions on data access and preparation, including how to access the raw data, preprocessed data, intermediate data, and generated data, etc.
        \item The authors should provide scripts to reproduce all experimental results for the new proposed method and baselines. If only a subset of experiments are reproducible, they should state which ones are omitted from the script and why.
        \item At submission time, to preserve anonymity, the authors should release anonymized versions (if applicable).
        \item Providing as much information as possible in supplemental material (appended to the paper) is recommended, but including URLs to data and code is permitted.
    \end{itemize}

\item {\bf Experimental setting/details}
    \item[] Question: Does the paper specify all the training and test details (e.g., data splits, hyperparameters, how they were chosen, type of optimizer) necessary to understand the results?
    \item[] Answer: \answerYes{}
    \item[] Justification: We provide anonymized code, configuration files, and instructions for dataset preparation and execution, together with the implementation details in Section~\ref{sec:experiments} and Appendix~\ref{app:settings}.
    \item[] Guidelines:
    \begin{itemize}
        \item The answer \answerNA{} means that the paper does not include experiments.
        \item The experimental setting should be presented in the core of the paper to a level of detail that is necessary to appreciate the results and make sense of them.
        \item The full details can be provided either with the code, in appendix, or as supplemental material.
    \end{itemize}

\item {\bf Experiment statistical significance}
    \item[] Question: Does the paper report error bars suitably and correctly defined or other appropriate information about the statistical significance of the experiments?
    \item[] Answer: \answerYes{}
    \item[] Justification: Main experimental results are reported as mean $\pm$ standard deviation over three independent runs with different random seeds, as stated in Sec.~4 and Table~\ref{tab:main_results}. This quantifies variability due to stochastic training and data sampling.

    \item[] Guidelines:
    \begin{itemize}
        \item The answer \answerNA{} means that the paper does not include experiments.
        \item The authors should answer \answerYes{} if the results are accompanied by error bars, confidence intervals, or statistical significance tests, at least for the experiments that support the main claims of the paper.
        \item The factors of variability that the error bars are capturing should be clearly stated (for example, train/test split, initialization, random drawing of some parameter, or overall run with given experimental conditions).
        \item The method for calculating the error bars should be explained (closed form formula, call to a library function, bootstrap, etc.)
        \item The assumptions made should be given (e.g., Normally distributed errors).
        \item It should be clear whether the error bar is the standard deviation or the standard error of the mean.
        \item It is OK to report 1-sigma error bars, but one should state it. The authors should preferably report a 2-sigma error bar than state that they have a 96\% CI, if the hypothesis of Normality of errors is not verified.
        \item For asymmetric distributions, the authors should be careful not to show in tables or figures symmetric error bars that would yield results that are out of range (e.g., negative error rates).
        \item If error bars are reported in tables or plots, the authors should explain in the text how they were calculated and reference the corresponding figures or tables in the text.
    \end{itemize}

\item {\bf Experiments compute resources}
    \item[] Question: For each experiment, does the paper provide sufficient information on the computer resources (type of compute workers, memory, time of execution) needed to reproduce the experiments?
    \item[] Answer: \answerYes{}
    \item[] Justification: The paper reports the hardware used for the experiments (single NVIDIA A100 GPU), and Appendix~\ref{app:complexity} provides runtime, GFLOPs, peak GPU memory, and storage overhead for representative runs.
    \item[] Guidelines:
    \begin{itemize}
        \item The answer \answerNA{} means that the paper does not include experiments.
        \item The paper should indicate the type of compute workers CPU or GPU, internal cluster, or cloud provider, including relevant memory and storage.
        \item The paper should provide the amount of compute required for each of the individual experimental runs as well as estimate the total compute. 
        \item The paper should disclose whether the full research project required more compute than the experiments reported in the paper (e.g., preliminary or failed experiments that didn't make it into the paper). 
    \end{itemize}
    
\item {\bf Code of ethics}
    \item[] Question: Does the research conducted in the paper conform, in every respect, with the NeurIPS Code of Ethics \url{https://neurips.cc/public/EthicsGuidelines}?
    \item[] Answer: \answerYes{}
    \item[] Justification: We have reviewed the NeurIPS Code of Ethics and, to the best of our knowledge, the research complies with it. The submission uses public datasets and pretrained models and does not involve deceptive, unsafe, or non-consensual experimental procedures.
    \item[] Guidelines:
    \begin{itemize}
        \item The answer \answerNA{} means that the authors have not reviewed the NeurIPS Code of Ethics.
        \item If the authors answer \answerNo, they should explain the special circumstances that require a deviation from the Code of Ethics.
        \item The authors should make sure to preserve anonymity (e.g., if there is a special consideration due to laws or regulations in their jurisdiction).
    \end{itemize}

\item {\bf Broader impacts}
    \item[] Question: Does the paper discuss both potential positive societal impacts and negative societal impacts of the work performed?
    \item[] Answer: \answerYes{}
    \item[] Justification: The paper includes a social impact section (Appendix~\ref{app:social_impact}) discussing both potential benefits and possible risks.
    \item[] Guidelines:
    \begin{itemize}
        \item The answer \answerNA{} means that there is no societal impact of the work performed.
        \item If the authors answer \answerNA{} or \answerNo, they should explain why their work has no societal impact or why the paper does not address societal impact.
        \item Examples of negative societal impacts include potential malicious or unintended uses (e.g., disinformation, generating fake profiles, surveillance), fairness considerations (e.g., deployment of technologies that could make decisions that unfairly impact specific groups), privacy considerations, and security considerations.
        \item The conference expects that many papers will be foundational research and not tied to particular applications, let alone deployments. However, if there is a direct path to any negative applications, the authors should point it out. For example, it is legitimate to point out that an improvement in the quality of generative models could be used to generate Deepfakes for disinformation. On the other hand, it is not needed to point out that a generic algorithm for optimizing neural networks could enable people to train models that generate Deepfakes faster.
        \item The authors should consider possible harms that could arise when the technology is being used as intended and functioning correctly, harms that could arise when the technology is being used as intended but gives incorrect results, and harms following from (intentional or unintentional) misuse of the technology.
        \item If there are negative societal impacts, the authors could also discuss possible mitigation strategies (e.g., gated release of models, providing defenses in addition to attacks, mechanisms for monitoring misuse, mechanisms to monitor how a system learns from feedback over time, improving the efficiency and accessibility of ML).
    \end{itemize}
    
\item {\bf Safeguards}
    \item[] Question: Does the paper describe safeguards that have been put in place for responsible release of data or models that have a high risk for misuse (e.g., pre-trained language models, image generators, or scraped datasets)?
    \item[] Answer: \answerNA{}
    \item[] Justification: The paper does not release high-risk generative models, scraped datasets, or other assets that would require special misuse safeguards. The method is an adaptation framework built on existing public vision-language models and benchmark datasets.
    \item[] Guidelines:
    \begin{itemize}
        \item The answer \answerNA{} means that the paper poses no such risks.
        \item Released models that have a high risk for misuse or dual-use should be released with necessary safeguards to allow for controlled use of the model, for example by requiring that users adhere to usage guidelines or restrictions to access the model or implementing safety filters. 
        \item Datasets that have been scraped from the Internet could pose safety risks. The authors should describe how they avoided releasing unsafe images.
        \item We recognize that providing effective safeguards is challenging, and many papers do not require this, but we encourage authors to take this into account and make a best faith effort.
    \end{itemize}

\item {\bf Licenses for existing assets}
    \item[] Question: Are the creators or original owners of assets (e.g., code, data, models), used in the paper, properly credited and are the license and terms of use explicitly mentioned and properly respected?
    \item[] Answer: \answerYes{}
    \item[] Justification: All data, models, and code in the paper respect the license.
    \item[] Guidelines:
    \begin{itemize}
        \item The answer \answerNA{} means that the paper does not use existing assets.
        \item The authors should cite the original paper that produced the code package or dataset.
        \item The authors should state which version of the asset is used and, if possible, include a URL.
        \item The name of the license (e.g., CC-BY 4.0) should be included for each asset.
        \item For scraped data from a particular source (e.g., website), the copyright and terms of service of that source should be provided.
        \item If assets are released, the license, copyright information, and terms of use in the package should be provided. For popular datasets, \url{paperswithcode.com/datasets} has curated licenses for some datasets. Their licensing guide can help determine the license of a dataset.
        \item For existing datasets that are re-packaged, both the original license and the license of the derived asset (if it has changed) should be provided.
        \item If this information is not available online, the authors are encouraged to reach out to the asset's creators.
    \end{itemize}

\item {\bf New assets}
    \item[] Question: Are new assets introduced in the paper well documented and is the documentation provided alongside the assets?
    \item[] Answer: \answerNA{}
    \item[] Justification: The paper does not release new assets.
    \item[] Guidelines:
    \begin{itemize}
        \item The answer \answerNA{} means that the paper does not release new assets.
        \item Researchers should communicate the details of the dataset\slash code\slash model as part of their submissions via structured templates. This includes details about training, license, limitations, etc. 
        \item The paper should discuss whether and how consent was obtained from people whose asset is used.
        \item At submission time, remember to anonymize your assets (if applicable). You can either create an anonymized URL or include an anonymized zip file.
    \end{itemize}

\item {\bf Crowdsourcing and research with human subjects}
    \item[] Question: For crowdsourcing experiments and research with human subjects, does the paper include the full text of instructions given to participants and screenshots, if applicable, as well as details about compensation (if any)? 
    \item[] Answer: \answerNA{}
    \item[] Justification: The paper does not involve crowdsourcing experiments or research with human subjects.
    \item[] Guidelines:
    \begin{itemize}
        \item The answer \answerNA{} means that the paper does not involve crowdsourcing nor research with human subjects.
        \item Including this information in the supplemental material is fine, but if the main contribution of the paper involves human subjects, then as much detail as possible should be included in the main paper. 
        \item According to the NeurIPS Code of Ethics, workers involved in data collection, curation, or other labor should be paid at least the minimum wage in the country of the data collector. 
    \end{itemize}

\item {\bf Institutional review board (IRB) approvals or equivalent for research with human subjects}
    \item[] Question: Does the paper describe potential risks incurred by study participants, whether such risks were disclosed to the subjects, and whether Institutional Review Board (IRB) approvals (or an equivalent approval/review based on the requirements of your country or institution) were obtained?
    \item[] Answer: \answerNA{}
    \item[] Justification: The paper does not involve human subjects research or crowdsourcing, so IRB approval or equivalent review was not required.
    \item[] Guidelines:
    \begin{itemize}
        \item The answer \answerNA{} means that the paper does not involve crowdsourcing nor research with human subjects.
        \item Depending on the country in which research is conducted, IRB approval (or equivalent) may be required for any human subjects research. If you obtained IRB approval, you should clearly state this in the paper. 
        \item We recognize that the procedures for this may vary significantly between institutions and locations, and we expect authors to adhere to the NeurIPS Code of Ethics and the guidelines for their institution. 
        \item For initial submissions, do not include any information that would break anonymity (if applicable), such as the institution conducting the review.
    \end{itemize}

\item {\bf Declaration of LLM usage}
    \item[] Question: Does the paper describe the usage of LLMs if it is an important, original, or non-standard component of the core methods in this research? Note that if the LLM is used only for writing, editing, or formatting purposes and does \emph{not} impact the core methodology, scientific rigor, or originality of the research, declaration is not required.
    \item[] Answer: \answerYes{}
    \item[] Justification: LLM usage is part of the method: Gemini 2.5 Pro is used offline to generate candidate class attributes for CTPA. Details, prompt template, and reproducibility notes are provided in Appendix~\ref{app:ctpa_llm}.
    \item[] Guidelines:
    \begin{itemize}
        \item The answer \answerNA{} means that the core method development in this research does not involve LLMs as any important, original, or non-standard components.
        \item Please refer to our LLM policy in the NeurIPS handbook for what should or should not be described.
    \end{itemize}

\end{enumerate}

\end{document}